\documentclass{article}

\PassOptionsToPackage{numbers, compress}{natbib}

\usepackage[preprint]{neurips_2023}

\usepackage{amsmath,amsfonts,bm}

\def\eqref#1{equation~\ref{#1}}

\def\1{\bm{1}}

\def\vo{{\bm{o}}}
\def\vp{{\bm{p}}}

\def\vx{{\bm{x}}}
\def\vy{{\bm{y}}}
\def\vz{{\bm{z}}}

\def\evp{{p}}

\def\evy{{y}}

\DeclareMathAlphabet{\mathsfit}{\encodingdefault}{\sfdefault}{m}{sl}
\SetMathAlphabet{\mathsfit}{bold}{\encodingdefault}{\sfdefault}{bx}{n}

\def\sN{{\mathbb{N}}}

\def\sR{{\mathbb{R}}}

\usepackage{hyperref}
\hypersetup{
    colorlinks=true,
    linkcolor=[rgb]{0.84,0.15,0.16},
    citecolor=[rgb]{0.12,0.47,0.71},
}

\usepackage{url}

\usepackage{times}
\usepackage{helvet}
\usepackage{courier}
\usepackage{nicefrac}       
\usepackage{microtype}      
\usepackage{xcolor}         
\usepackage{color,colortbl}
\usepackage{caption}
\definecolor{myred}{rgb}{0.84,0.15,0.16}
\definecolor{myblue}{rgb}{0.12,0.47,0.71}

\usepackage{enumitem}

\usepackage{graphicx}
\usepackage{caption}
\usepackage{subcaption}
\usepackage{wrapfig}

\usepackage{booktabs}       
\usepackage{tabularx}
\usepackage{array}
\usepackage{multirow}
\usepackage{siunitx}
\usepackage{arydshln}
\usepackage{amssymb} 
\usepackage{dsfont} 

\usepackage{amsthm} 

\usepackage{algorithm}
\usepackage{algpseudocode}

\usepackage{soul}

\title{Asymptotic Midpoint Mixup\\for Margin Balancing and Moderate Broadening}

\author{%
  Hoyong Kim, Semi Lee, Kangil Kim\thanks{corresponding author}\\
  Artificial Intelligence Graduate School \\
  Gwangju Institute of Science and Technology  \\
  Gwangju 61005, Republic of Korea \\
  \texttt{hoyong.kim.21@gm.gist.ac.kr, \{leesemi0719, kangil.kim.01\}@gmail.com} \\
}

\begin{document}

\maketitle

\begin{abstract}
   In the feature space, the collapse between features invokes critical problems in representation learning by remaining the features undistinguished.
   Interpolation-based augmentation methods such as mixup have shown their effectiveness in relieving the collapse problem between different classes, called inter-class collapse.
   However, intra-class collapse raised in coarse-to-fine transfer learning has not been discussed in the augmentation approach. 
   To address them, we propose a better feature augmentation method, $\textit{asymptotic midpoint mixup}$. 
   The method generates augmented features by interpolation but gradually moves them toward the midpoint of inter-class feature pairs. 
   As a result, the method induces two effects: 1) balancing the margin for all classes and 2) only moderately broadening the margin until it holds maximal confidence.
   We empirically analyze the collapse effects by measuring \textit{alignment} and \textit{uniformity} with visualizing representations.
   Then, we validate the intra-class collapse effects in coarse-to-fine transfer learning and the inter-class collapse effects in imbalanced learning on long-tailed datasets. 
   In both tasks, our method shows better performance than other augmentation methods.
\end{abstract}

%
%
\section{Introduction}
\label{sec:intro}

Feature augmentation in neural networks has been effective in regularization by handling margins in feature space~\cite{manifold_mixup}. 
The approach generates a \textit{feature}, which indicates a hidden representation of a layer created from an input, and its confidence information from involved original features.
A similar approach in the perspective of directly repositioning features, \textit{contrastive learning}, has also benefited from the feature control ~\cite{contrast_chen2020simple, contrast_he2020momentum}. 
Contrastive learning methods learn features distant from a decision boundary by getting centroids of classes further away from each other and gathering positive pairs closer. 
This kind of training strategy decreases intra-class feature distance while increasing inter-class feature distance, measured by \textit{alignment} and \textit{uniformity}, respectively. 

\begin{figure*}[h!]
    \centering
    \includegraphics[width=0.98\linewidth]{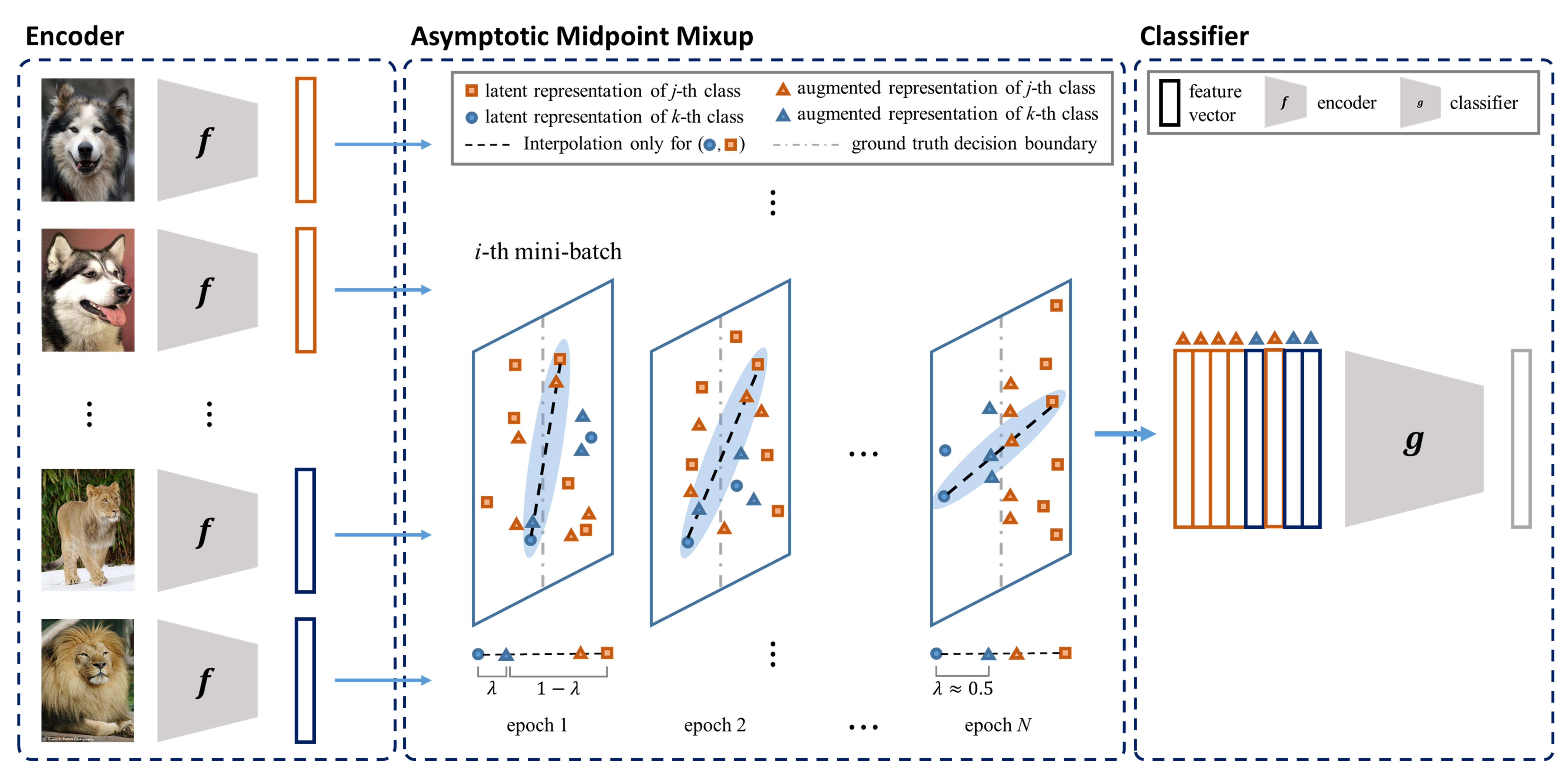}
    \caption{
    Overview of Asymptotic Midpoint Mixup. 
    (left) Feature vectors of input samples came from the pre-trained encoder. 
    (middle) Asymptotic midpoint mixup generates augmented features and their labels based on interpolation. 
    Examples for understanding the interpolation are highlighted as cyan. 
    The ratio between two different features is controlled by $\lambda$ and this parameter has asymptotically decreased from 1.0 to 0.5 until the end of training. 
    The augmented features are created as a mini-batch size at the same rate. 
    (right) Finally, the augmented features are passed to the classifier.
    }
    \label{fig:ammixup_overview}
\end{figure*}

Despite the strength of contrastive loss, two problems have been recently discussed: collapse of intra-class and inter-class features~\cite{contrast_li2022targeted, contrast_chen2022perfectly}. 
The first problem has been reported in coarse-to-fine transfer learning, where all features are closely located on the centroids of each class as the alignment excessively decreases~\cite{contrast_chen2022perfectly}. 
The second problem has been introduced in supervised contrastive learning (SupCon~\cite{contrast_supcon_khosla2020supervised}), which uses ground truth labels to create positive and negative pairs.
However, it causes unbalanced margins on long-tailed datasets by overwhelming numerical dominance of the head classes, and this decreases the image classification performance on them.
In contrast, interpolation-based augmentation methods, especially mixup~\cite{mixup}, have improved the performance of models on long-tailed datasets and contrastive self-supervised learning~\cite{mixup_ssl} but the collapse problems have yet to be deeply analyzed.

In this motivation, we studied inter-class and intra-class collpase effects on the interpolation-based augmentation methods, especially mixup and manifold mixup, in the coarse-to-fine transfer learning and imbalanced learning.
The same method as SupCon was used, and as a result, we found that mixup and manifold mixup also suffer from similar collapse problems.
To address these collapse problems, we propose a better interpolation-based augmentation method, $\textit{asymptotic midpoint mixup}$ (AM-mixup). 
This method generates augmented features by elaborately controling the ratio of interpolation. 
To be more detail, the augmented features asymptotically approach to the midpoint between two classes as the model trained, as shown in~\autoref{fig:ammixup_overview}.
As a result, the margins of the classes become balanced and moderately broad while weakening the collapse effects.

In an experiment on a toy task, we validate the effect of collapses by using alignment and uniformity metrics for mixup, manifold mixup, and our method.
We empirically verify the impact of our method in comparison with the feature augmentation methods in image classification tasks on long-trailed and coarse-to-fine transfer datasets.

In summary, our main contributions are three-fold:
\begin{itemize}
    \item We address the issues of inter-class and intra-class collapse in feature augmentation approaches and show their impact by analyzing alignment and uniformity.
    \item We propose a better interpolation-based augmentation method, \textit{asymptotic midpoint mixup}, to address the collapse problems by balancing and moderately broadening the margin in the feature space.
    \item We empirically analyze the effects and performance of our method and other feature augmentation methods in imbalanced learning and coarse-to-fine transfer learning.
\end{itemize}

%
%
\section{Background}
\paragraph{Intra-class collapse.}
Contrastive loss causes features in positive pairs to be close and they become invariant to some noise factors. 
In other words, the encoder is forced to ensure that similar samples must be placed in similar locations in the feature space. 
However, the features tend to cluster at one point due to the exorbitant attraction between positive pairs.
This phenomenon limits the expressiveness of the model, and it is particularly critical in some environments such as coarse-to-fine transfer learning. 
More specifically, if a model is pretrained with coarse-grained labels and then fine-tuned with fine-grained labels, the model is unlikely to classify input samples to fine-grained labels due to the collapsed features. 
In particular, features in the same superclass are prone to collapse on the centroids of the superclass in supervised contrastive learning even though the fine-grained classifier tries to classify them as their own subclasses. 
We called this problem as \textit{intra-class collapse}. 
To measure intra-class collapse, \textit{intra-class alignment} has been proposed, which represents the closeness of positive pairs~\cite{measure_wang2020understanding, contrast_li2022targeted}. 
The intra-class alignment can be measured by following:
\begin{equation}
\begin{array}{c}
{\bf A} = \frac{1}{C} \sum_{i=1}^C \frac{1}{\mid \textrm{F}_{i} \mid^{2}} \sum_{\bm{v}_{j}, \bm{v}_{k} \in \textrm{F}_{i}} \lVert \bm{v}_{j} - \bm{v}_{k} \rVert_{2} 
\end{array}
\label{eq:alignment}
\end{equation}
where $C$ is the number of classes, $\bm{v}$ is a feature vector, and $\textrm{F}_{i}$ is the set of features from class $i$.
$|\cdot|$ and $\lVert\cdot\rVert_2$ means cardinality and L2-norm, respectively.

\paragraph{Inter-class collapse}
Contrastive learning achieves high performance thanks to the property that the centroids of the classes move further away from each other due to repulsion between anchors and negative samples. 
However, when dealing with long-tailed datasets, supervised contrastive learning may suffer from feature collapsing across difference classes.
In this case, the model may prioritize maximizing the distance between head classes to reduce the loss. 
For this reason, the contrastive loss is not evenly distributed among all classes. 
In this situation, features in the tail classes would be collapsed to each other. 
We refer to this collapse as \textit{inter-class collapse}, which hinders the model's ability to learn a regular simplex of features. 
This is critical in imbalanced learning. 
The inter-class collapse can be quantified using \textit{inter-class} and \textit{neighborhood uniformity}, metrics that prioritize the uniform distribution of representations on the unit hypersphere~\cite{measure_wang2020understanding, contrast_chen2020simple}. 
The inter-class uniformity assesses the distance between different centroids of classes, while the neighborhood uniformity examines the convergence of the tail classes.
These two types of metrics can be measured using the following $\mathbf{U}$ and $\mathbf{U}_k$, respectively:
\begin{equation}
\begin{array}{c}
{\bf U} = \frac{1}{C(C-1)} \sum_{i=1}^C \sum_{j=1, j \neq i}^C \lVert \bar{\bm{v}}_{i} - \bar{\bm{v}}_{j}\rVert_{2}
\end{array}
\label{eq:uniformity_inter_class}
\end{equation}
\begin{equation}
\begin{array}{c}
{\bf U}_{k} = \frac{1}{C} \sum_{i=1}^C \underset{j_{1}, \cdots, j_{k}}{\min}(\sum_{l=1}^k \lVert \bar{\bm{v}}_{i} - \bar{\bm{v}}_{j_{l}}\rVert_{2})
\end{array}
\label{eq:uniformity_neighbor}
\end{equation}
where $\bar{\bm{v}}_{i}$ is the center of samples from class $i$ on the hypersphere: $\bar{\bm{v}}_{i} = \frac{\sum_{\bm{v}_{j} \in \textrm{F}_{i}} \bm{v}_{j}}{\lVert \sum_{\bm{v}_{j} \in \textrm{F}_{i}} \bm{v}_{j} \rVert_{2}}$.

%
%

\section{Motivation}
\label{sec:motivation}
In this section, we first present our motivation for preliminary experiments on alignment and uniformity for an interpolation-based method - mixup and manifold mixup. 
To quantitatively measure the intra-class and inter-class collapse, we examine intra-class alignment in coarse-to-fine transfer learning and inter-class uniformity and $k$=1 neighborhood uniformity in imbalanced learning, respectively.

\paragraph{General Settings.}
To visualize features on two dimensional space, we utilized CNN-Vis2D to have two dimensional feature vectors as the output of the encoder.
We trained it with the softmax function, cross entropy, and SGD with momentum 0.9 and weight decay 5e-4. 
The initial learning rate was set to 0.1 and divided to 0.2 at epoch 30, 60, and 80 during 100 epochs. 
We also set the mixup alpha to 1.0.

\begin{figure}[t]
    \centering
    \begin{subfigure}[b]{0.245\linewidth}
        \includegraphics[width=\linewidth]{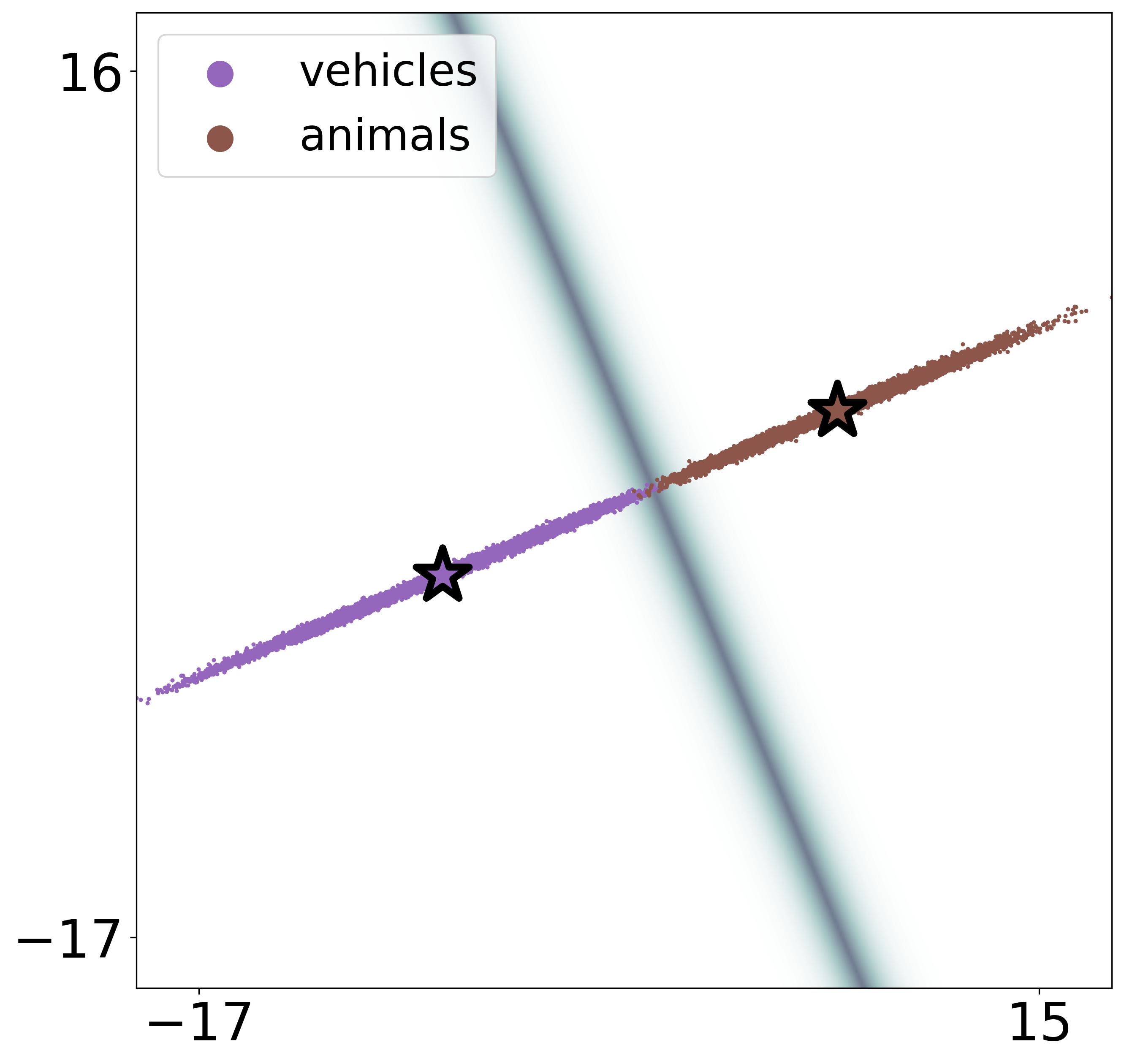}
        \captionsetup{justification=centering}
        \caption{CE\\in Coarse\\~}
        \label{subfig:default_coarse}
    \end{subfigure}
    \begin{subfigure}[b]{0.24\linewidth}
        \includegraphics[width=\linewidth]{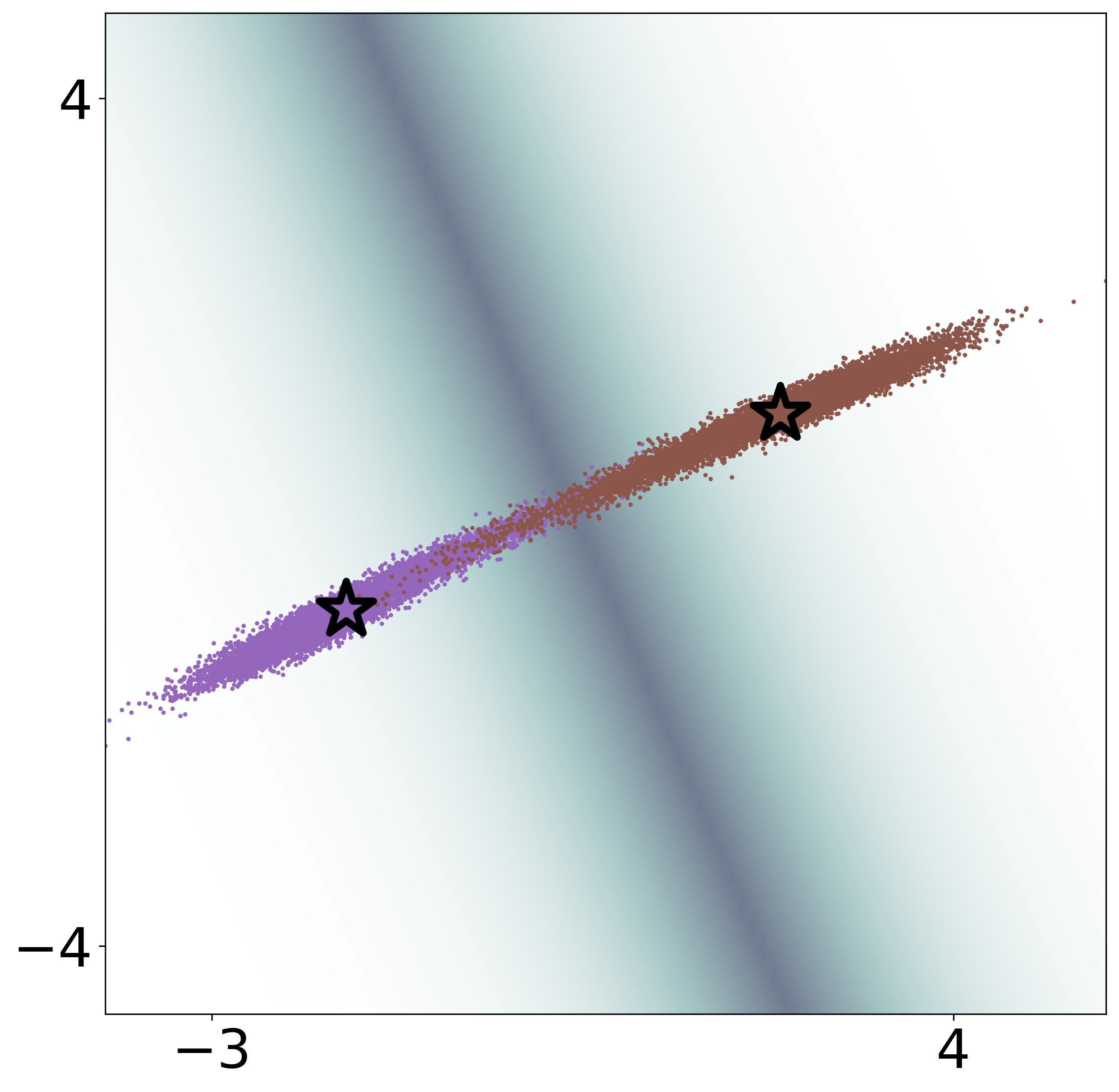}
        \captionsetup{justification=centering}
        \caption{Mixup\\in Coarse\\~}
        \label{subfig:mixup_coarse}
    \end{subfigure}
    \begin{subfigure}[b]{0.24\linewidth}
        \includegraphics[width=\linewidth]{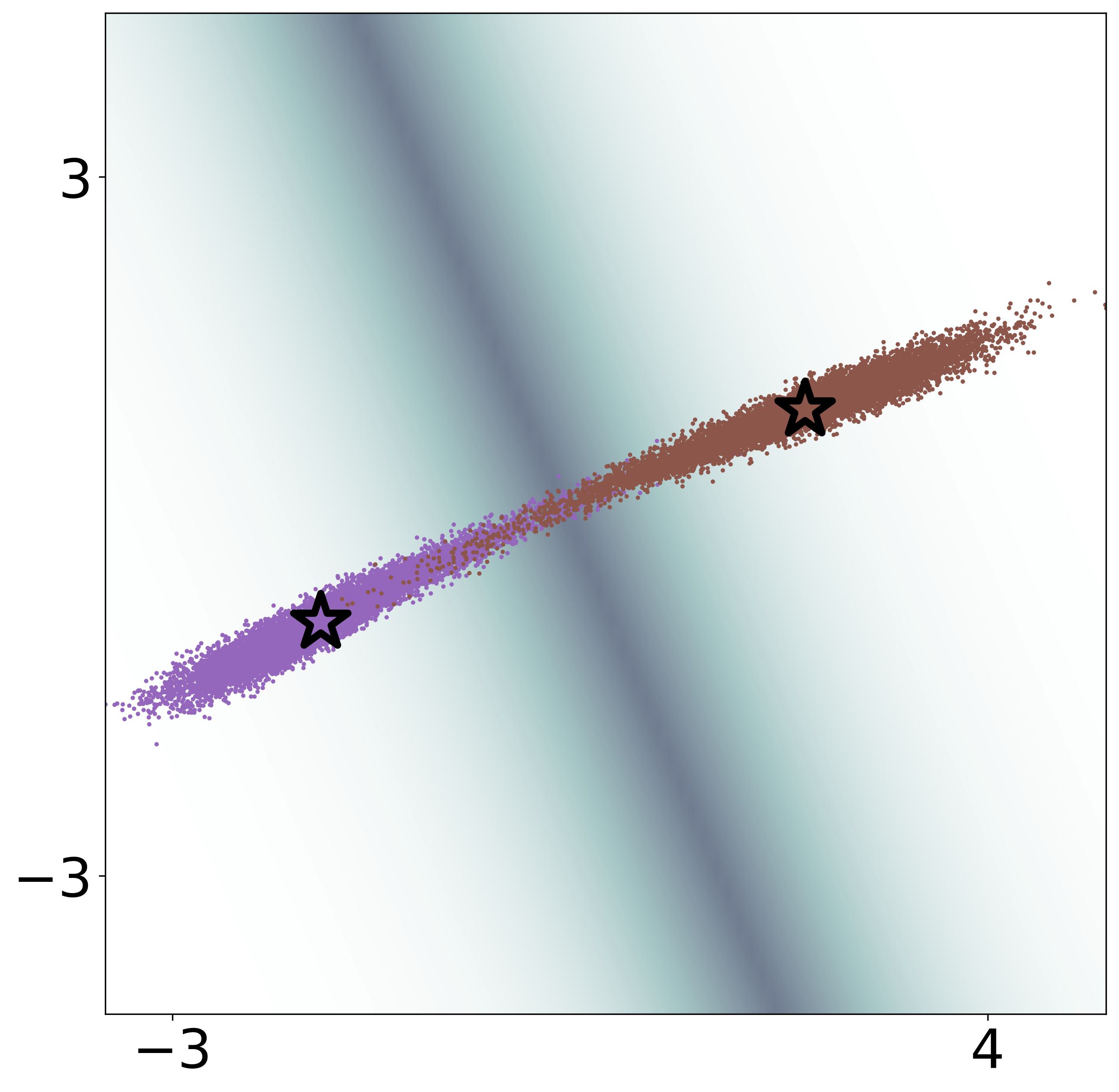}
        \captionsetup{justification=centering}
        \caption{Manifold Mixup\\in Coarse\\~}
        \label{subfig:manifold_coarse}
    \end{subfigure}
    \begin{subfigure}[b]{0.24\linewidth}
        \includegraphics[width=\linewidth]{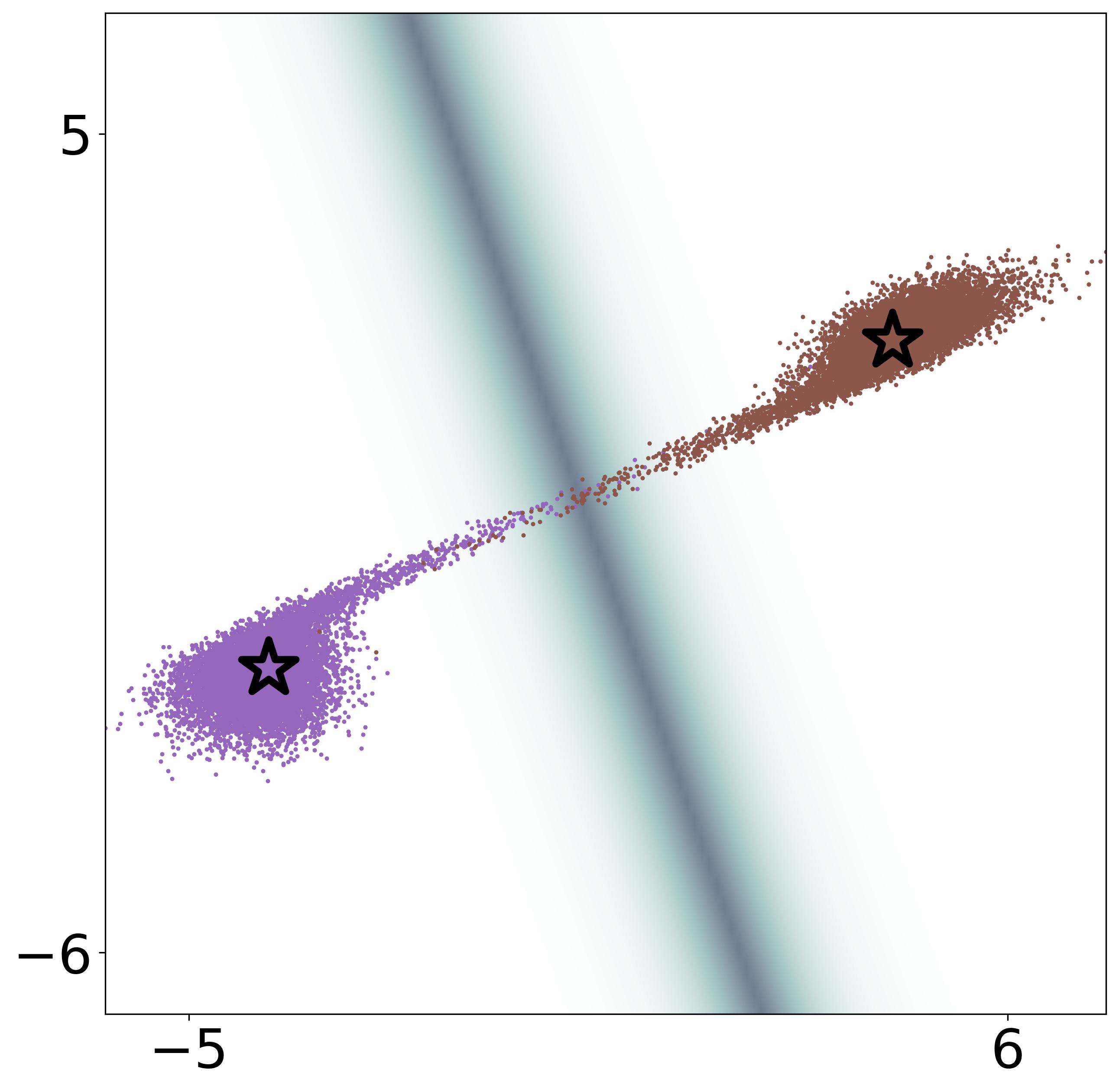}
        \captionsetup{justification=centering}
        \caption{AM-mixup\\in Coarse\\~}
        \label{subfig:ama_coarse}
    \end{subfigure}
    \begin{subfigure}[b]{0.245\linewidth}
        \includegraphics[width=\linewidth]{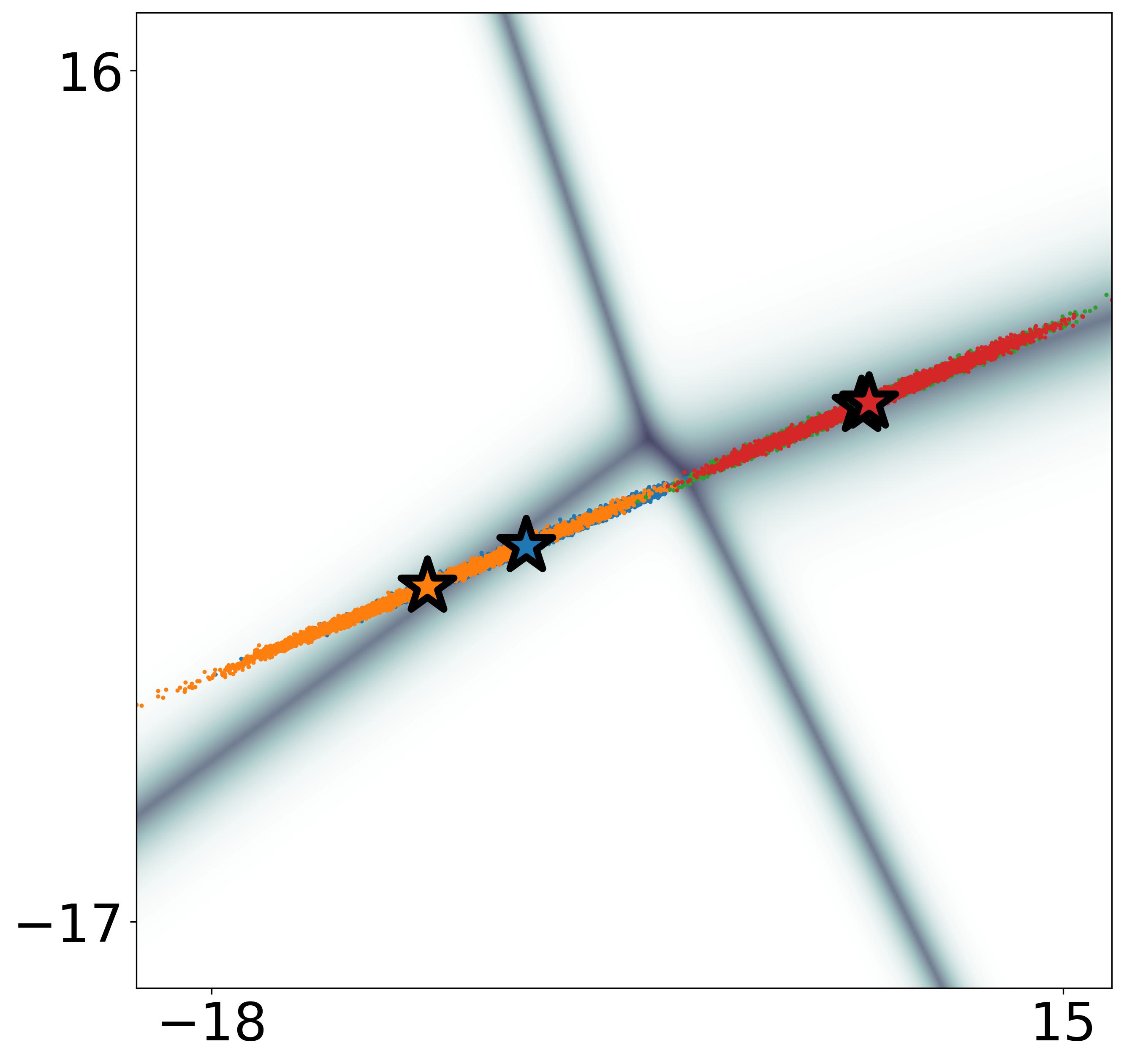}
        \captionsetup{justification=centering}
        \caption{CE in Fine\\~}
        \label{subfig:default_fine}
    \end{subfigure}
    \begin{subfigure}[b]{0.24\linewidth}
        \includegraphics[width=\linewidth]{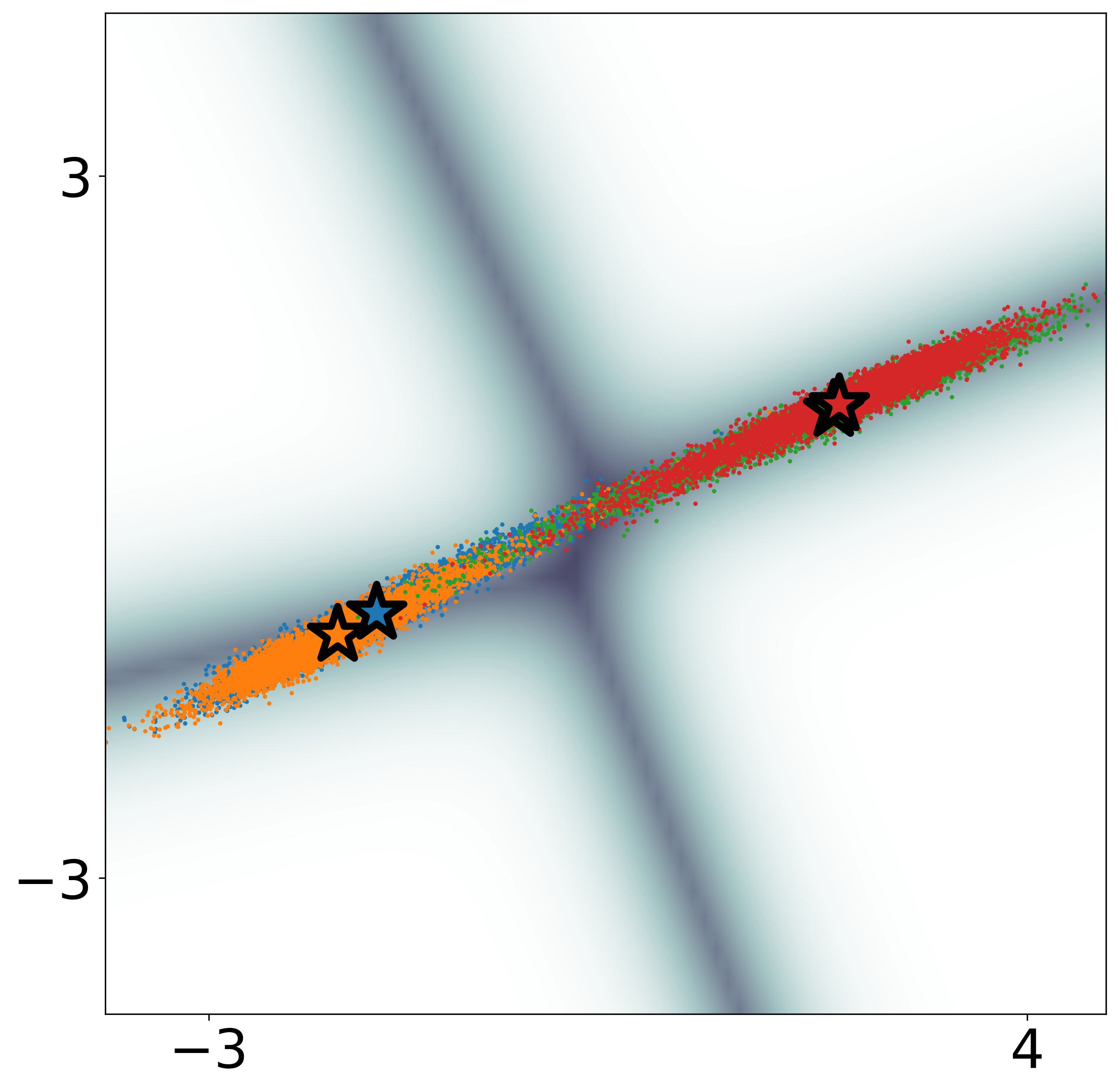}
        \captionsetup{justification=centering}
        \caption{Mixup in Fine\\~}
        \label{subfig:mixup_fine}
    \end{subfigure}
    \begin{subfigure}[b]{0.24\linewidth}
        \includegraphics[width=\linewidth]{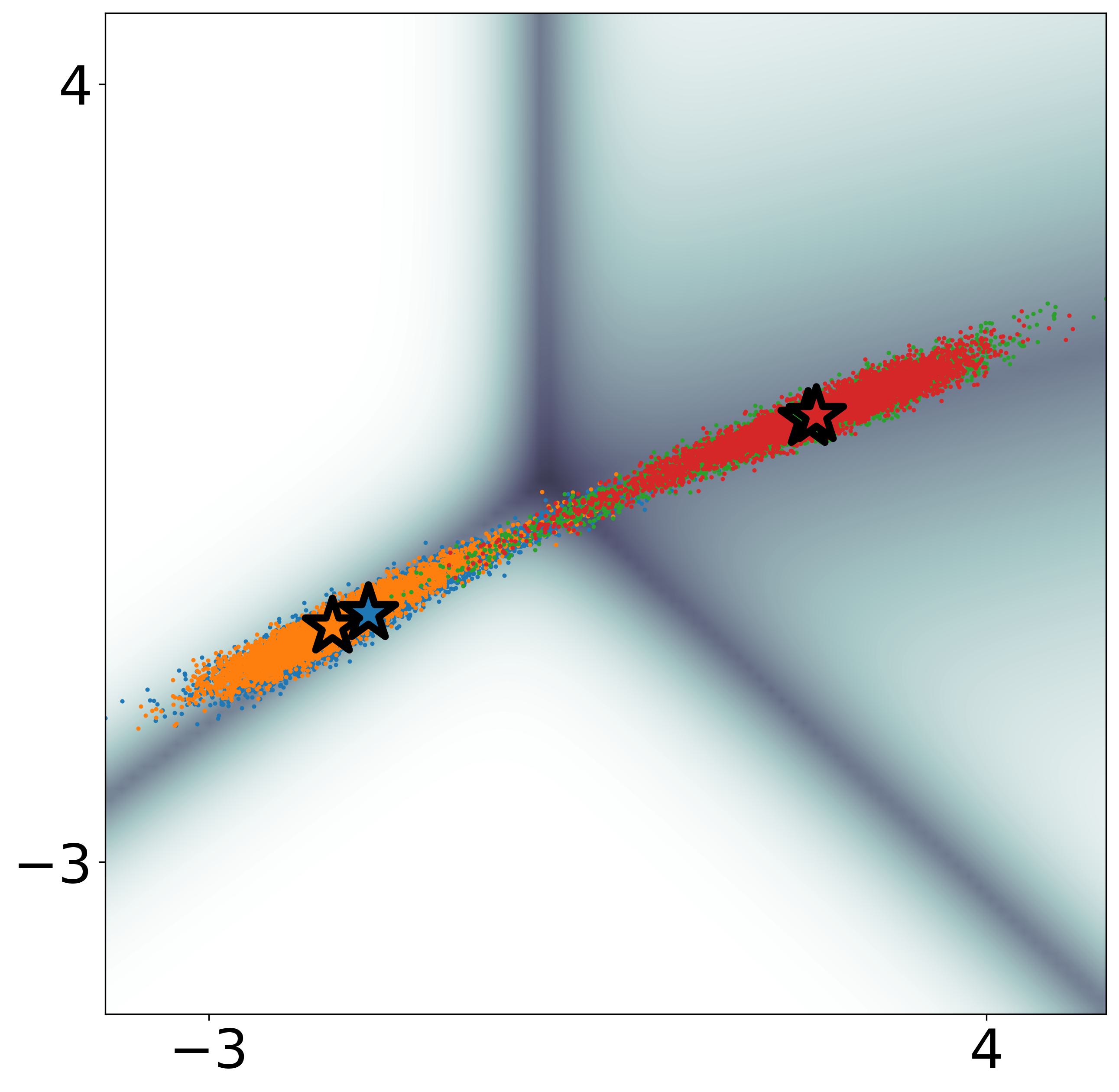}
        \captionsetup{justification=centering}
        \caption{Manifold Mixup\\in Fine}
        \label{subfig:manifold_fine}
    \end{subfigure}
    \begin{subfigure}[b]{0.24\linewidth}
        \includegraphics[width=\linewidth]{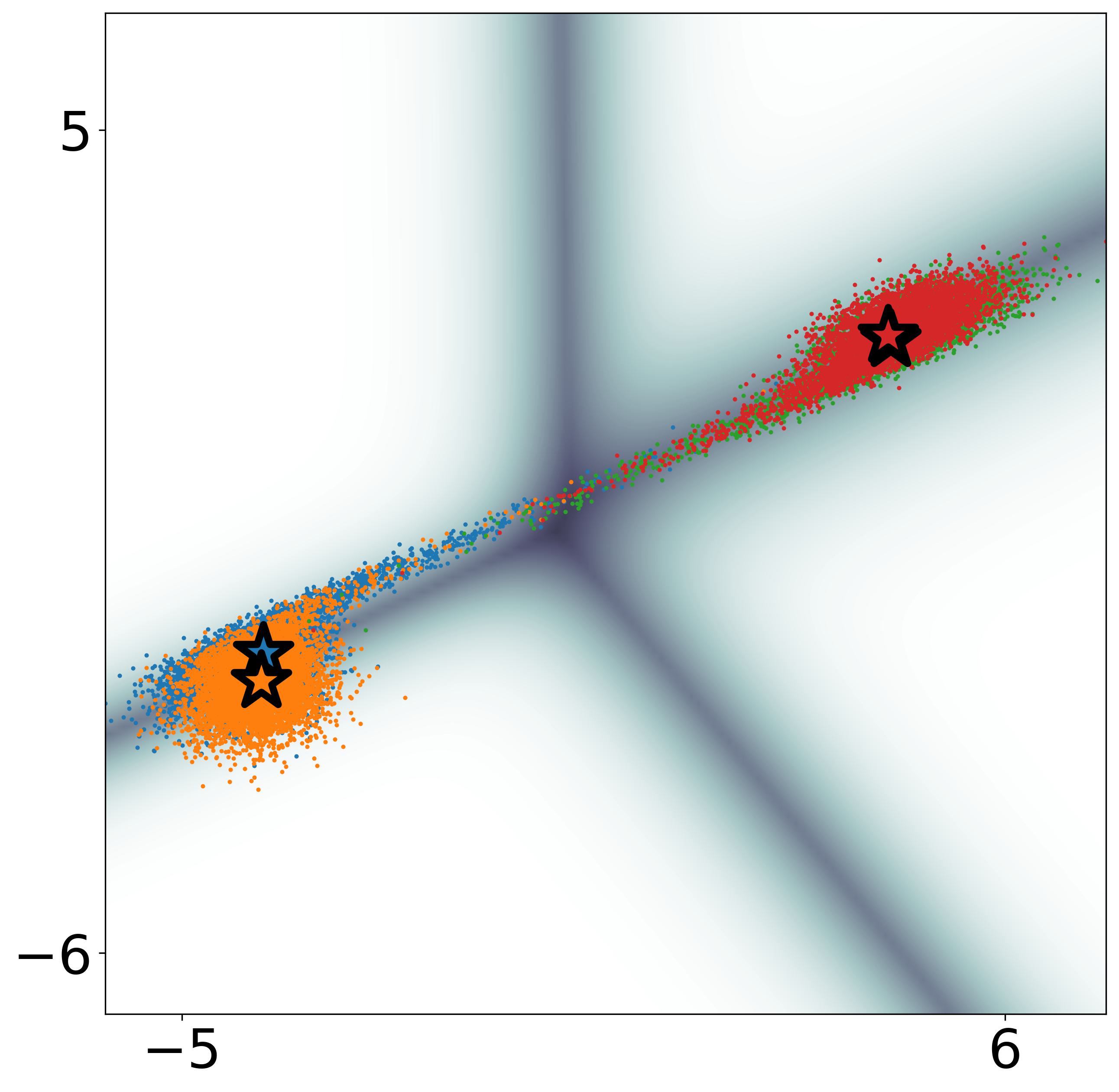}
        \captionsetup{justification=centering}
        \caption{AM-mixup in Fine\\~}
        \label{subfig:ama_fine}
    \end{subfigure}
\caption{
Comparison results on Mini-CIFAR-Coarse and Mini-CIFAR-Fine. 
In the case of coarse-to-fine transfer learning, features have various scales when trained on the coarse-grained datasets, and this diversity of the scale is helpful for fine-tuning on the fine-grained datasets. 
Our method shows better spread in the feature space than mixup and manifold mixup but not enough to CE.
The background shows the confidence landscape as a heatmap, where the lighter the color, the higher the confidence (min: 0.0, max: 1.0). 
Features are colored according to the class to which they belong. 
(CE: Cross entropy without any augmentation method, AM-mixup: our method)
}
\label{fig:motivation_coarse2fine}
\end{figure}

\begin{table}[h!]
\small
  \centering
  \caption{
  Results of Coarse-to-Fine Transfer Learning on Mini-CIFAR.
  \textbf{A} refers to the alignment. 
  Generally, lower alignment indicates better performance, but this is not the case in coarse-to-fine transfer learning. 
  Thus, $\Downarrow$ signifies that low alignment is better, but too low is not optimal. 
  Best in bold and second best in underline. 
  }
  \begin{tabular}{@{}lcc@{}}
    \toprule
    Method & \textbf{A} $\Downarrow$ & Test Acc.(\%) \\
    \midrule
    CE & 3.7535 & \textbf{61.08} \\
    mixup & \underline{0.9460} & 58.43 \\
    manifold mixup & \textbf{0.9185} & 55.73 \\
    AM-mixup & 0.9923 & \underline{60.77} \\
    \bottomrule
  \end{tabular}
  \label{tab:toy_coarse_to_fine}
\end{table}

\paragraph{Coarse-to-Fine Transfer Learning on Mini-CIFAR-Coarse.}
For the coarse-to-fine transfer learning, we utilized Mini-CIFAR dataset derived from the CIFAR10 dataset.
This involved selecting only 20K samples from the first four classes of CIFAR10 dataset: airplane, automobile, bird, and cat.
Finally, Mini-CIFAR-Coarse consists of two superclasses: vehicles (airplane, automobile) and animals (bird, cat). 
In coarse-to-fine transfer learning, the model's encoder is first trained on Mini-CIFAR-Coarse.
Then, the pretrained encoder was frozen, and the classifier is fine-tuned on Mini-CIFAR, which consists of four subclasses as labels. 
The key objective of coarse-to-fine transfer learning is to optimize the feature distribution during the training of the coarse-grained dataset and separate them into subclasses during fine-tuning.
In terms of this point, we hypothesize that higher alignment in the feature space results in superior model performance.
As shown in \autoref{tab:toy_coarse_to_fine}, we have quantitatively verified this assumption by observing that CE, which has the highest alignment, outperforms the other three methods. 
We observe a similar tendency in the other three methods. 
As a result, we conclude that our method can assist in alleviating intra-class collapse during mixup.
This conclusion is supported by both empirical and qualitative evidence presented in \autoref{fig:motivation_coarse2fine}.

\begin{figure}[t]
    \centering
    \begin{subfigure}[b]{0.245\linewidth}
        \includegraphics[width=\linewidth]{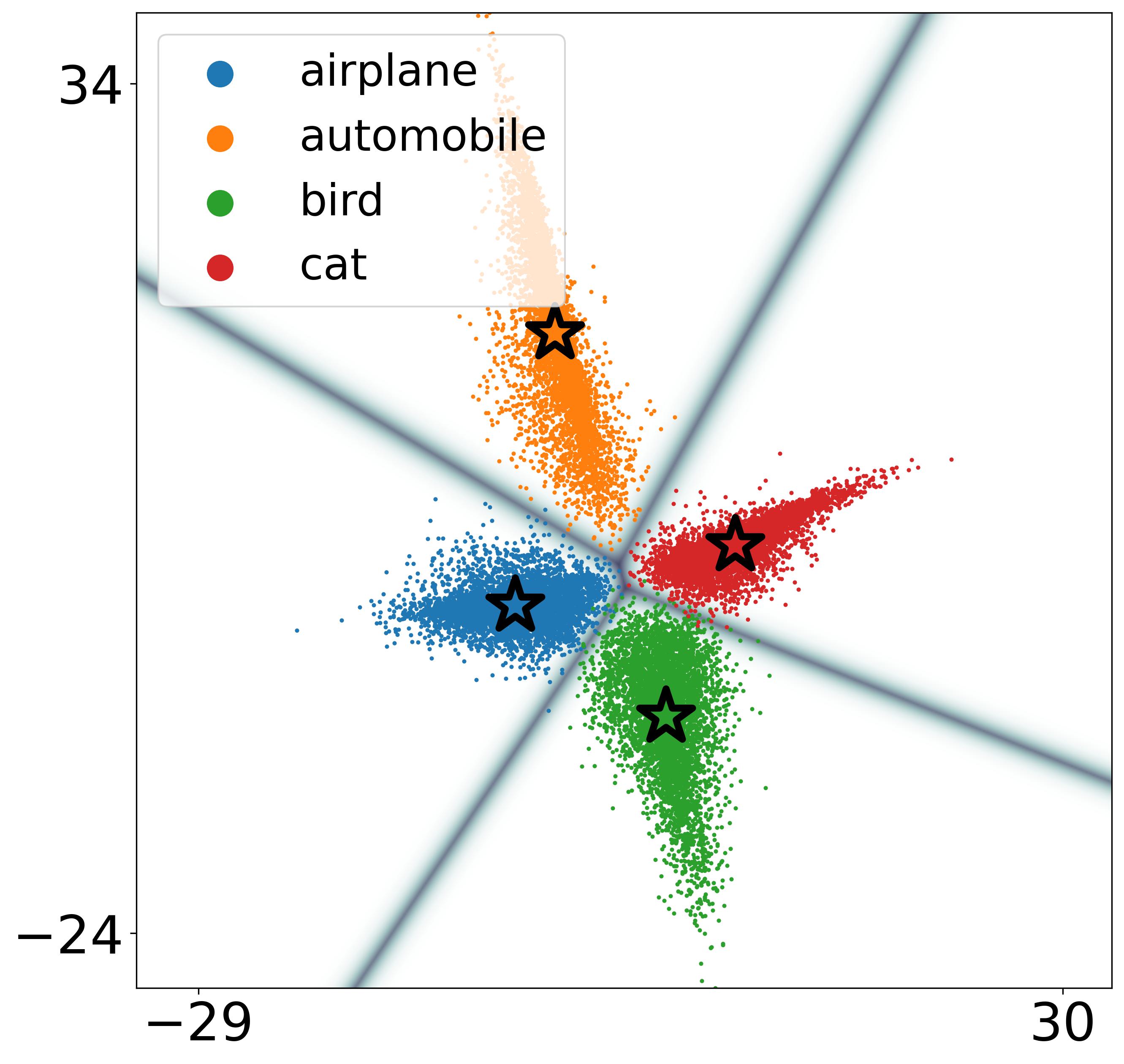}
        \captionsetup{justification=centering}
        \caption{CE\\~}
        \label{subfig:default_balanced}
    \end{subfigure}
    \begin{subfigure}[b]{0.24\linewidth}
        \includegraphics[width=\linewidth]{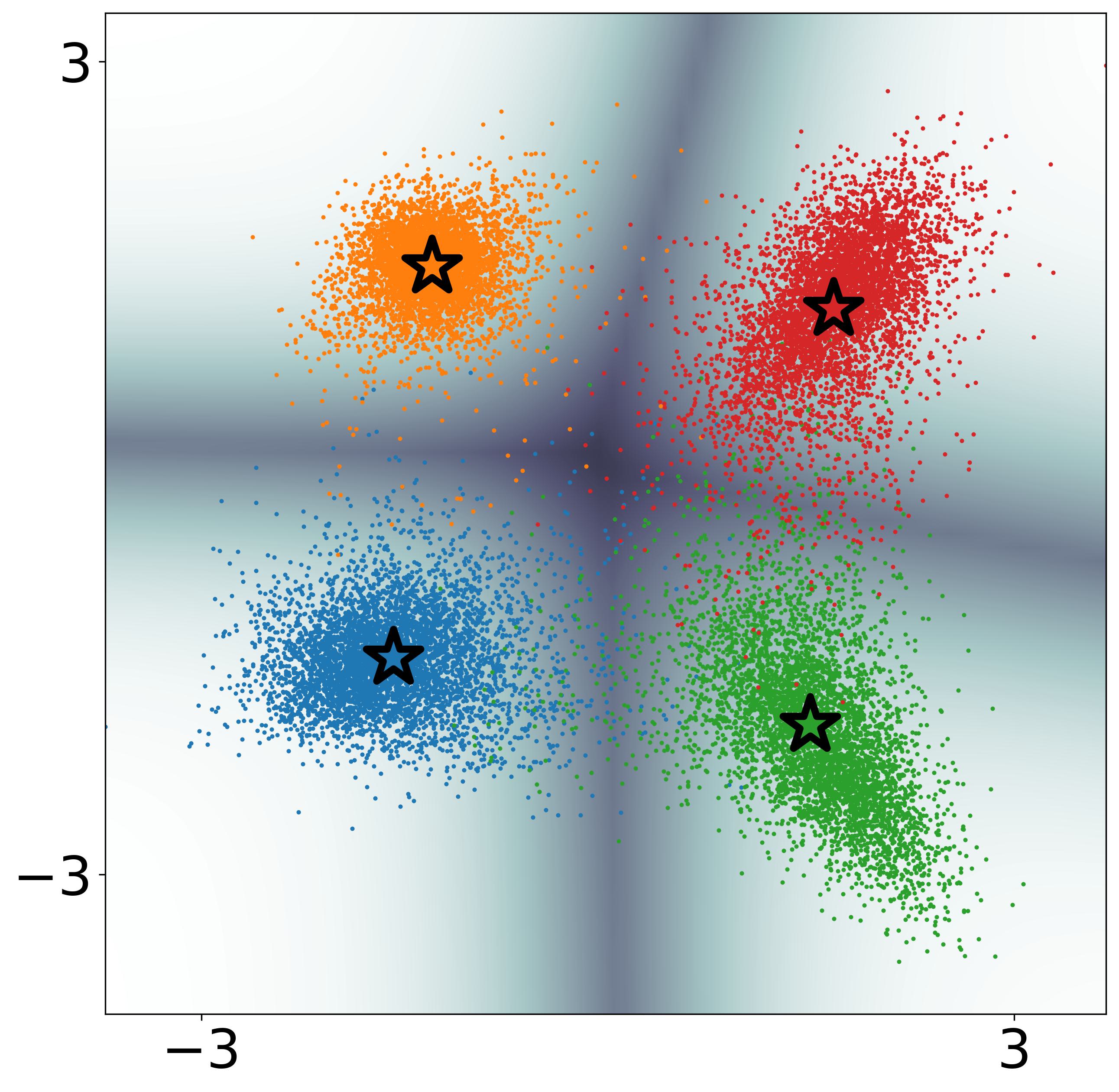}
        \captionsetup{justification=centering}
        \caption{Mixup\\~}
        \label{subfig:mixup_balanced}
    \end{subfigure}
    \begin{subfigure}[b]{0.24\linewidth}
        \includegraphics[width=\linewidth]{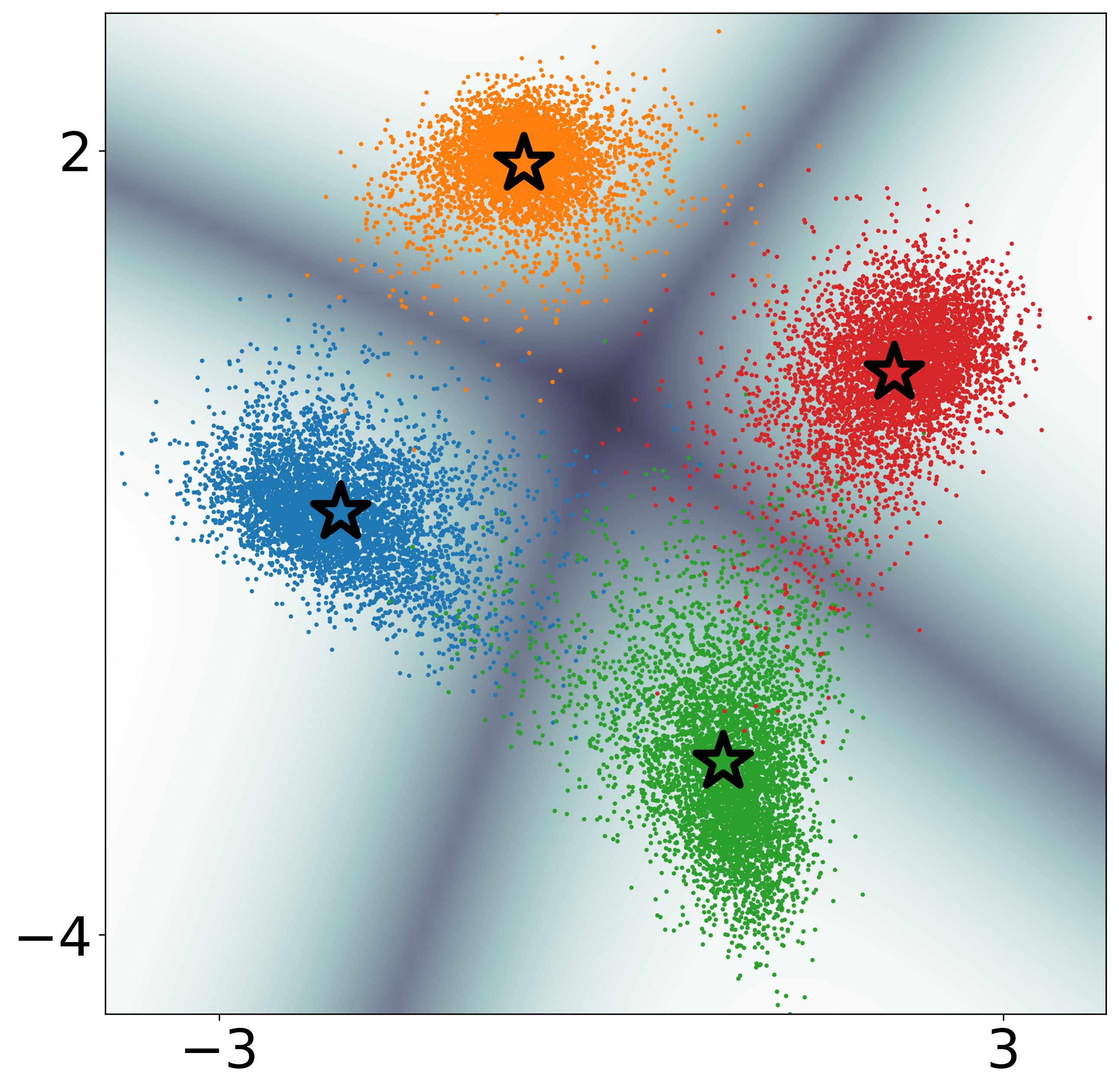}
        \captionsetup{justification=centering}
        \caption{Manifold Mixup\\~}
        \label{subfig:manifold_balanced}
    \end{subfigure}
    \begin{subfigure}[b]{0.24\linewidth}
        \includegraphics[width=\linewidth]{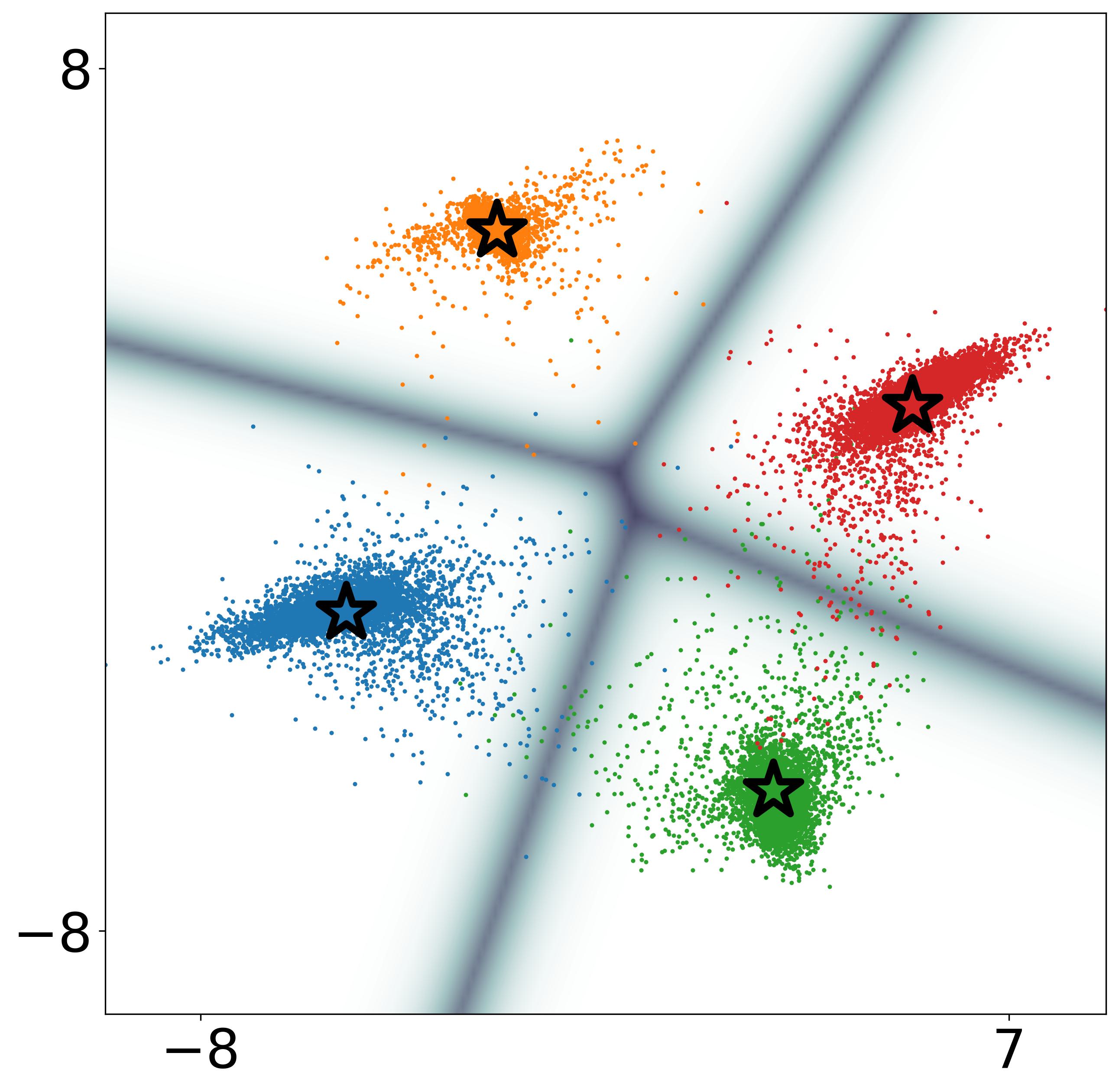}
        \captionsetup{justification=centering}
        \caption{AM-mixup\\~}
        \label{subfig:ama_balanced}
    \end{subfigure}
    \begin{subfigure}[b]{0.24\linewidth}
        \includegraphics[width=\linewidth]{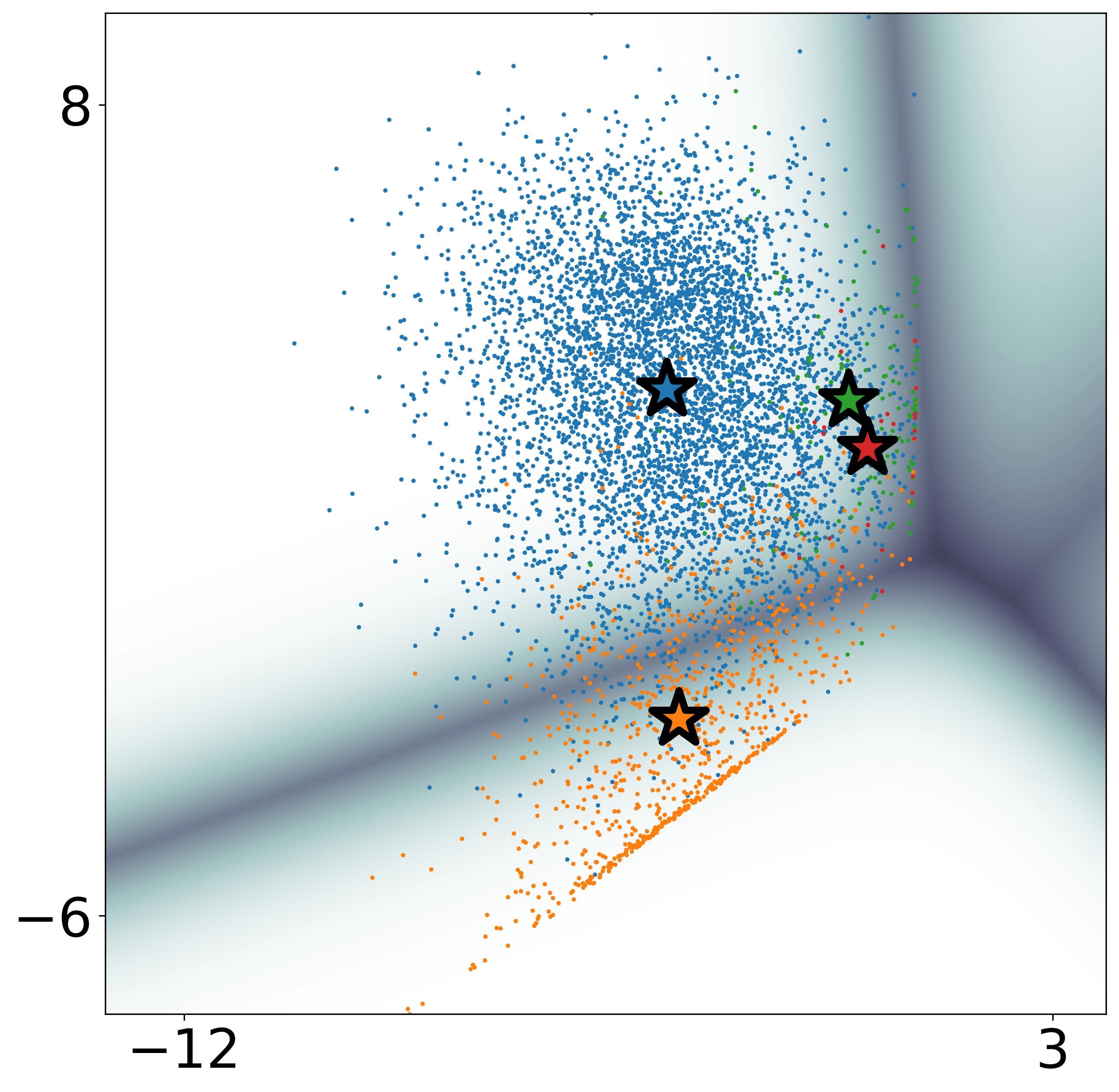}
        \captionsetup{justification=centering}
        \caption{CE in LT}
        \label{subfig:default_lt}
    \end{subfigure}
    \begin{subfigure}[b]{0.24\linewidth}
        \includegraphics[width=\linewidth]{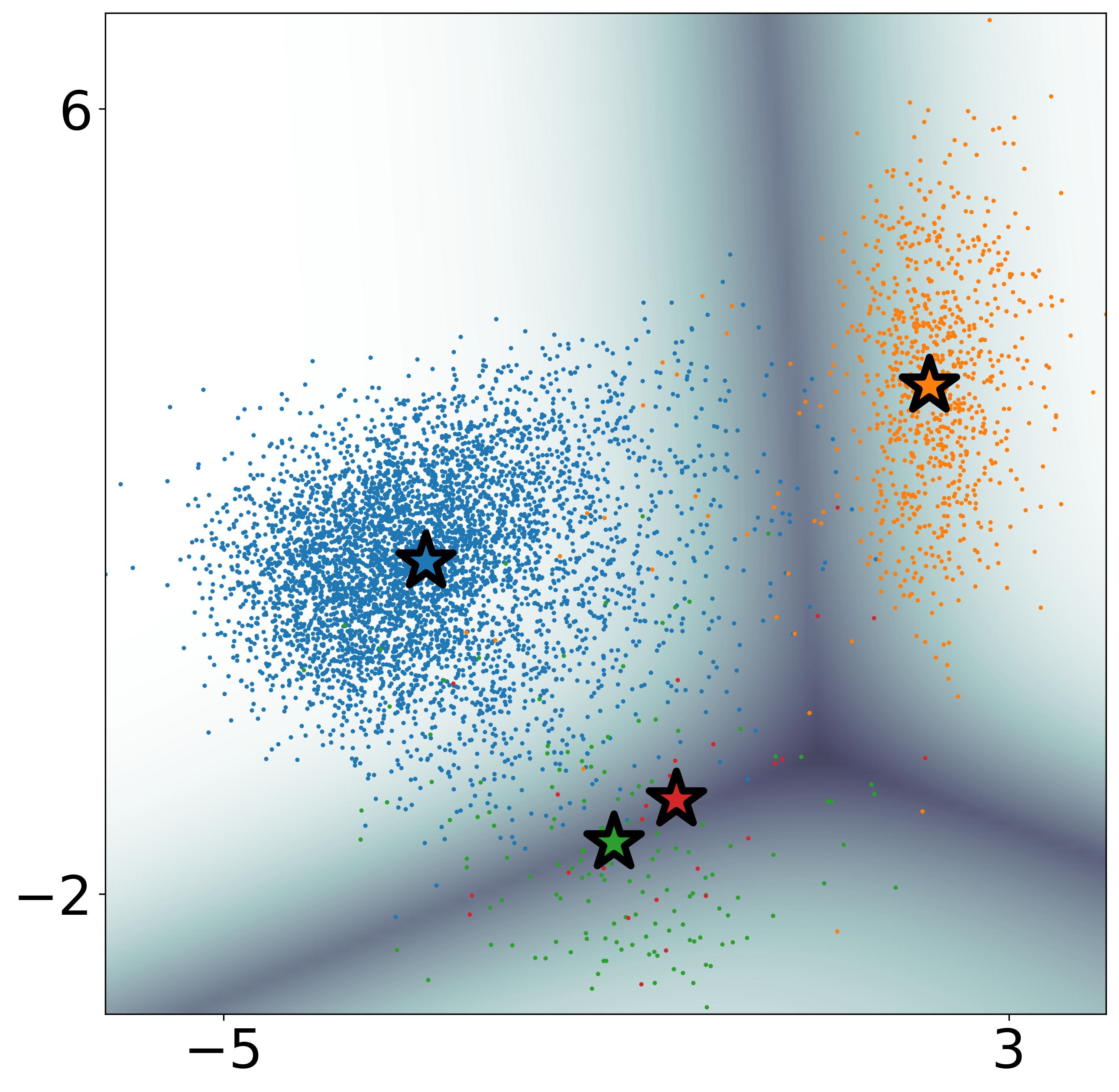}
        \captionsetup{justification=centering}
        \caption{Mixup in LT}
        \label{subfig:mixup_lt}
    \end{subfigure}
    \begin{subfigure}[b]{0.24\linewidth}
        \includegraphics[width=\linewidth]{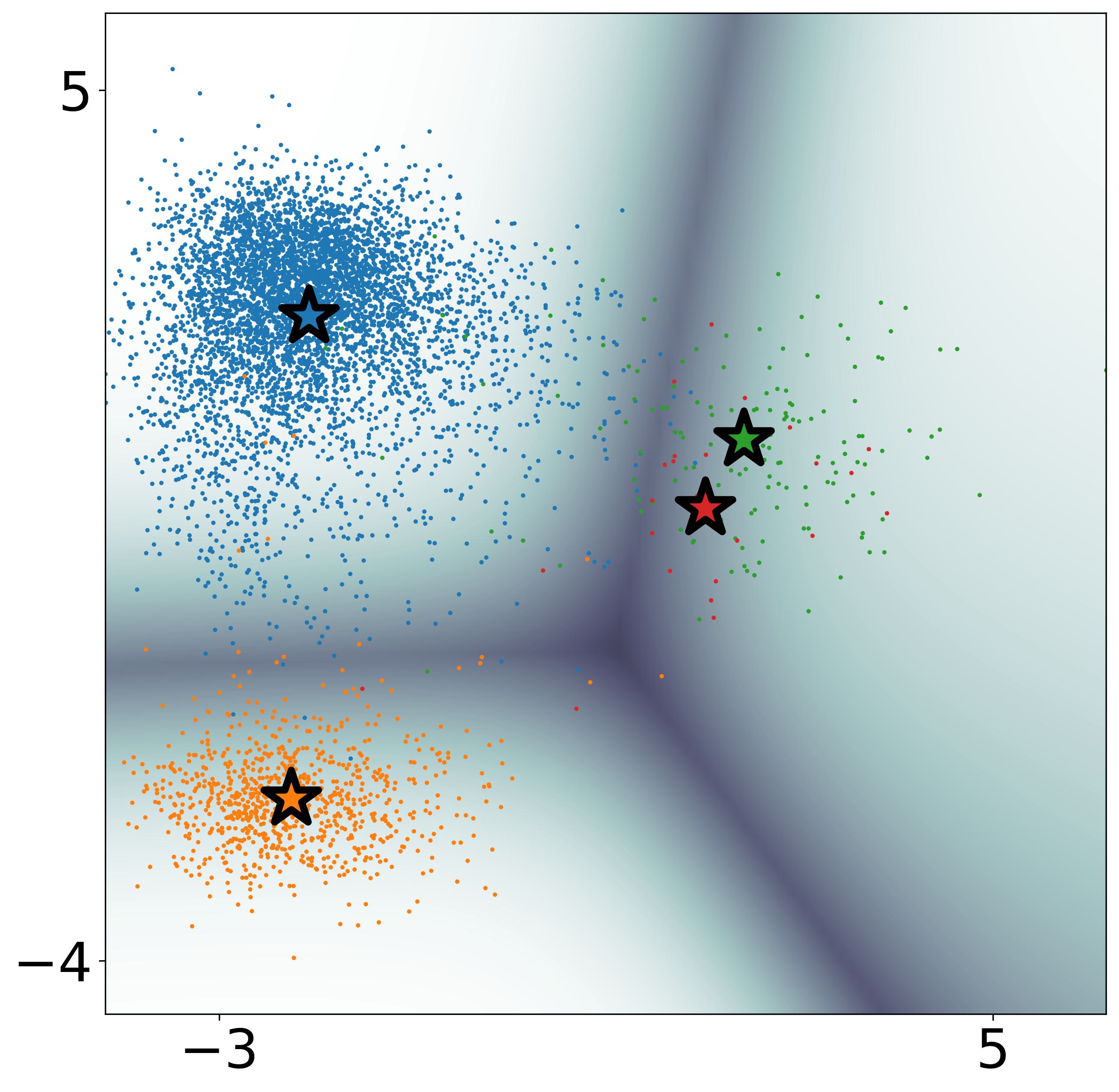}
        \captionsetup{justification=centering}
        \caption{Manifold Mixup in LT}
        \label{subfig:manifold_lt}
    \end{subfigure}
    \begin{subfigure}[b]{0.24\linewidth}
        \includegraphics[width=\linewidth]{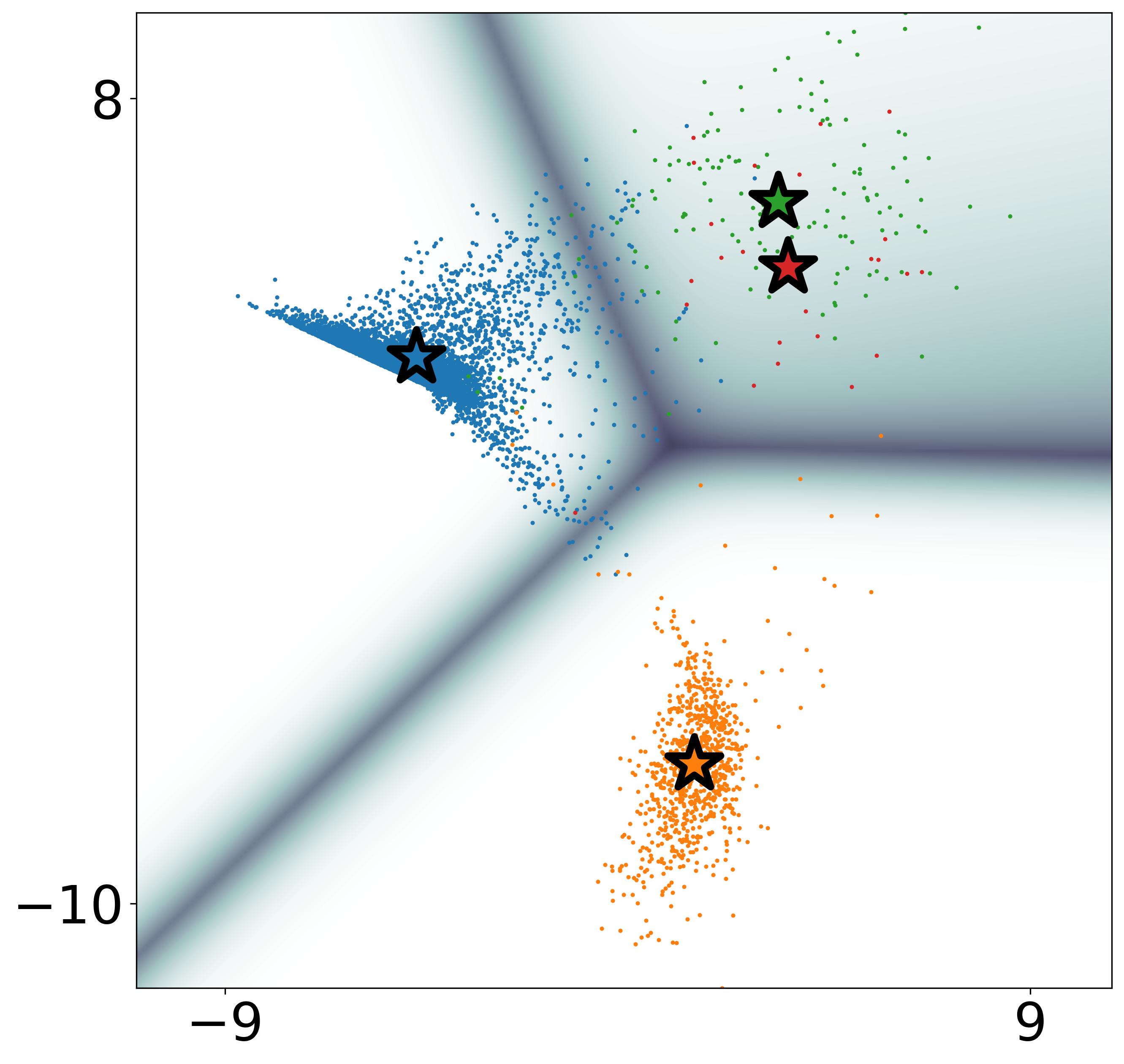}
        \captionsetup{justification=centering}
        \caption{AM-mixup in LT}
        \label{subfig:ama_lt}
    \end{subfigure}
\caption{
Comparison results on Mini-CIFAR and Mini-CIFAR-LT.
In a normal environment, CE shows that the features have an extremely large scale, while the scale of the features in mixup and manifold mixup is very small. 
In between, our method shows moderately broadening the margins of classes than others.
In imbalanced learning, the centroids of the tail classes (bird and cat) are very close to the decision boundary in mixup and manifold mixup.
CE shows worse results as classifying input samples in the tail classes to the head class (airplane).
The background shows the confidence landscape as a heatmap, where the lighter the color, the higher the confidence (min: 0.0, max: 1.0). 
Features are colored according to the class to which they belong. 
(CE: Cross entropy without any augmentation method, AM-mixup: our method)
}
\label{fig:motivation_lt}
\end{figure}

\begin{table}[h!]
\centering
\caption{Results of Imbalanced Learning on Mini-CIFAR-LT (imb=200). Best in bold. $\uparrow$ means a high value is better. }
    \begin{subtable}[h]{0.47\linewidth}
        \scriptsize
        \centering
        \caption{\textbf{U} and \textbf{U$_{1}$} means inter-class uniformity and neighborhood($k$=1) uniformity, respectively}
        \begin{tabular}{@{}lccc@{}}
        \toprule
        Method & \textbf{U} $\uparrow$ & \textbf{U}$_{1}$ $\uparrow$ & Test Acc.(\%) \\
        \midrule
        CE & 1.0353 & 0.5226 & 42.52 \\
        mixup & 1.4198 & 0.8271 & 59.40 \\
        manifold mixup & \textbf{1.4623} & 0.8532 & 65.42 \\
        AM-mixup & 1.4477 & \textbf{0.9024} & \textbf{66.90} \\
        \bottomrule
        \end{tabular}
        \label{subtab:toy_lt}
    \end{subtable}
    \hfill
    \begin{subtable}[h]{0.47\linewidth} 
        \scriptsize
        \centering
        \caption{Detailed comparison results of each part (Accuracy (\%))}
        \begin{tabular}{@{}lcccc@{}}
        \toprule
        Method & Many & Median & Few & All \\
        \midrule
        CE & 98.10 & 67.50 & 2.25 & 42.52 \\
        mixup & \textbf{99.00} & 95.00 & 21.80 & 59.40 \\
        manifold mixup & 98.90 & \textbf{96.30} & 33.25 & 65.42 \\
        AM-mixup & 97.00 & 96.00 & \textbf{37.30} & \textbf{66.90} \\
        \bottomrule
        \end{tabular}
        \label{subtab:toy_lt_detail}
    \end{subtable}
\label{tab:toy_lt}
\end{table}

\paragraph{Imbalanced Learning on Mini-CIFAR-LT.}
For the imbalanced learning, we set the Mini-CIFAR-LT dataset from CIFAR10.
Mini-CIFAR-LT consists of four classes as labels, the same as Mini-CIFAR, but the imbalance factor is set to 200.
This means that the number of samples $n_{max}$ in the head class is 200 times greater than that of the tail class $n_{min}$, i.e., $\frac{n_{max}}{n_{min}}=200$. 
In imbalanced learning, we train the model on the Mini-CIFAR-LT and obtain accuracy on three splits of classes: \{\textit{Head-Many} (more than 1,000 images): [airplane], \textit{Median} (200 to 1,000 images): [automobile], \textit{Tail-Few} (less than 200 images): [bird, cat]\}.
The key objective of imbalanced learning concerns the distance between centroids and the balance of margins.
With regard to this focus, not only is inter-class uniformity significant, but neighborhood uniformity also holds value in obtaining superior inductive bias in imbalanced learning.
Consequently, we hypothesize that a model will perform better in imbalanced learning when these factors are high, especially neighborhood uniformity, because there are few classes in Mini-CIFAR-LT.
As shown in Table~\ref{subtab:toy_lt}, we have quantitatively validated this assumption by finding that CE exhibits the lowest inter-class uniformity and neighborhood uniformity, resulting in the method's weakest overall performance among the three. 
While manifold mixup displays higher inter-class uniformity than our method, it is reversed for neighborhood uniformity, where AM-mixup shows greater test accuracy than manifold mixup. 
Table~\ref{subtab:toy_lt_detail} more effectively illustrates the impact of adjusting the margin balance.
Each category was calculated as the ratio of the number of test samples in the classes.
The higher precision in the \textit{few} category indicates that the model achieved a more balanced margin in those instances.
Therefore, we conclude that our method can be helpful to achieve a balanced margin in imbalanced learning.
This claim is supported by empirical and qualitative evidence in \autoref{fig:motivation_lt}.

\paragraph{Experimental Results and Analysis}
As shown in the preliminary experiments conducted on Mini-CIFAR, AM-mixup increases the margin more widely than mixup and manifold mixup. 
As a result, we found that AM-mixup makes the model train moderately broad margins and balancing them by asymptotically moving the augmented features to midpoint in the feature space.
The effectiveness of AM-mixup was shown empirically and qualitatively in~\autoref{fig:motivation_coarse2fine} and~\autoref{fig:motivation_lt} and quantitatively in~\autoref{tab:toy_coarse_to_fine} and~\autoref{tab:toy_lt}.
In the following, we introduce a better interpolation method, \textit{asymptotic midpoint mixup} and how it invokes moderately broad and balanced margins by adjusting the mixup rate and creating the augmented features and labels accordingly.

%
%
\paragraph{Notations.}
Let $\mathbf{D}=\{(\vx_i, c_i)|1 \le i \le n, i \in \sN\}$ be the set of pairs of an input vector and its label where $\vx_i\in \sR^d$ and $c_i \in C$ for the class index set $C$ and the pair index $i$. 
We define $\vy_i=[\evy_1, \evy_2, ... , \evy_{|C|}] \in \sR^{|C|}$ as one-hot encoding vector for $c_i$, where $\evy_{c_i}=1$. 
The feature vector and output vector of $i$-th input sample $\vx_i$ are notated as $\vz_i$ and $\vo_i$, respectively.
The confidence $\vp$ comes from $\sigma(\vo)$, where $\sigma(\cdot)$ is a function that normalizes an input vector into a range that leads to probabilistic interpretations, similarly to softmax. 
In this paper, we used softmax function for $\sigma(\cdot)$.

\subsection{Preliminary}
In mixup~\cite{mixup}, input sample pairs are randomly selected and mixed up each other as below:

\begin{equation}
\begin{array}{c}
    \vx^{(i,j)} = \lambda \cdot \vx_{i} + (1-\lambda) \cdot \vx_{j} \\ [3pt]
    \vy^{(i,j)} = \lambda \cdot \vy_{i} + (1-\lambda) \cdot \vy_{j} \\     
\end{array}
\label{eq:virtual_samples}
\end{equation}
where $\vx_i$ and $\vx_j$ are \textit{i}-th and \textit{j}-th input samples chosen in the mini-batch, respectively.
$\vx^{(i,j)}$ is the mixed input sample, which is linearly interpolated by the mixup rate $\lambda\thicksim$ Beta$(\alpha, \alpha)$, where mixup alpha $\alpha$ is a hyperparameter.
In the same way, the mixed label vector $\vy^{(i,j)}$ is made by linearly interpolating the \textit{i}-th input sample's label vector $\vy_i$ and the \textit{j}-th input sample's label vector $\vy_j$.

\subsection{Proposed Method}
\paragraph{Interpolation-Based Feature Generation and One-sided Labeling.}
In AM-mixup, augmented features and labels are created as
\begin{equation}
\begin{array}{c}
    \vz^{(i,j)} = \lambda_{am} \cdot \vz_{i} + (1-\lambda_{am}) \cdot \vz_{j} \\ [3pt]
    c^{(i,j)} =
    \begin{cases}
        c_i, & \mbox{if }\lambda_{am}\ge0.5 \\
        c_j, & \mbox{if }\lambda_{am}<0.5
    \end{cases}
\end{array}
\label{eq:virtual_samples}
\end{equation}
where $\vz^{(i,j)}$ is an augmented feature created by interpolating $\vz_{i}$ and $\vz_{j}$ randomly selected from the mini-batch, and the resulting label is now referred to as $c^{(i,j)}$ and called a \textit{one-sided label}.
This process takes place in the feature space, and the one-sided labels are determined by adjusting a parameter $\lambda_{am}$ to asymptotically move them closer to the decision boundary. 
Unlike other interpolation-based methods, the labels are unambiguously determined on one side, leading to increased margin for the corresponding class.

\paragraph{Asymptotic Move of Augmented Features.}
Confidence is an important factor in estimating the decision boundary. 
However, relying on the one-sided labels as the ground truth in early training is an untrustworthy method because neural networks can make erroneous predictions.
To reduce this risk, we use a scheduler that uses the training accuracy to adjust $\lambda_{am}$ with greater sensitivity, as illustrated in Eq.~\ref{eq:alpha_control_tr_acc}.
\begin{equation}
    \lambda_{am} = f(v_{acc}) = e^{-\beta \cdot v_{acc}}
\label{eq:alpha_control_tr_acc}
\end{equation}
where $v_{acc} \in [0, 1]$ means the real value of training accuracy at each epoch. 
$\beta$ is a hyperparameter to decide how $\lambda_{am}$ decreases as the training accuracy. 
$\beta$ should be set to 0.67, resulting in an exponential decrease of lambda from 1.0 to approximately 0.5. 
As the mixup rate approaches 0.5, augmented features result in a significant loss and cause the model to train with a large margin.
Generally, such large loss increases the margin between a feature and decision boundary by maximizing feature confidence.
However, after obtaining maximal confidence, original features stop moving further away from the boundary.
Due to near-zero gradients, the distance between intra-class features and their centroids is moderately maintained without excessive convergence pressure.

Despite this advantage, a mixup rate of about 0.5 appears to render the training of the model difficult during training's later stages.
To alleviate this phenomenon, we experimentally tested various $\beta$ values and determined that the model performs best when $\beta=0.34$.

\paragraph{Training Loss for Augmented Features.}
The loss functions of cross entropy, mixup and AM-mixup can be expressed in the formula as below: 

\begin{equation}
    \mathcal{L_{\textrm{CE}}}(\vx, \vy) =  -\sum_{k=1}^C \evy_k \log \evp_k 
    \label{eq:loss_ce}
\end{equation}
\begin{equation}
    \begin{aligned}
    \mathcal{L_{\textrm{mixup}}}(\vx^{(i,j)}, \vy^{(i,j)}) = \lambda &\cdot \mathcal{L_{\textrm{CE}}}(\vx^{(i,j)}, \vy_{i}) \\
    + (1 - \lambda) &\cdot \mathcal{L_{\textrm{CE}}}(\vx^{(i,j)}, \vy_{j})    
    \end{aligned}
    \label{eq:loss_mixup}
\end{equation}
\begin{equation}
    \mathcal{L_{\textrm{am}}}(\vx^{(i,j)}, \vy^{(i,j)})
    = -\sum_{k=1}^C \evy_k^{(i,j)} \log \evp_k^{(i,j)} 
    \label{eq:loss_ama}
\end{equation}
As illustrated in Eqs.~\ref{eq:loss_ce},~\ref{eq:loss_mixup}, and~\ref{eq:loss_ama}, AM-mixup loss is reminiscent of cross entropy loss, contrary to mixup loss, which employs both labels.
AM-mixup loss solely utilize one side wherein $\lambda_{am}$ is larger than the other.
Through this one-sided labeling, AM-mixup can expand the margin more extensively than mixup by attempting to increase the confidence of augmented features towards those one-sided labels.

%
%

\begin{table*}[t]
\tiny
\centering
\caption{Results of Imbalanced Learning on CIFAR10/100-LT. All experiments were conducted using 20 trials with random seeds. (Accuracy (\%): \textit{mean}$_{\textit{std}}$)}
\begin{tabular}{@{}llcrrrrrrrr@{}}
    \toprule
     & \multirow{2}{*}{Method} & \multirow{2}{*}{ref} & \multicolumn{4}{c}{CIFAR10-LT} & \multicolumn{4}{c}{CIFAR100-LT} \\ 
    \cmidrule(ll){4-7} \cmidrule(ll){8-11}
    & & & Many & Median & Few & All & Many & Median & Few & All\\
    \midrule
    \multirow{8}{*}{\rotatebox[origin=c]{90}{imb factor: 100}} 
    & CE & \cite{mislas_zhong2021improving}  
    & - & - & - & 70.40~~~~~~~ 
    & - & - & - & 38.40~~~~~~~ \\
    & mixup & \cite{mislas_zhong2021improving}
    & - & - & - & 73.10~~~~~~~ 
    & - & - & - & 39.60~~~~~~~  \\
    & manifold mixup & \cite{bbn_zhou2020bbn} 
    & - & - & - & 72.96~~~~~~~ 
    & - & - & - & 38.25~~~~~~~ \\
    \cmidrule(lr){2-11}
    & CE$^{\dag}$ & - 
    & 91.99$_{3.02}$ & 71.72$_{1.67}$ & 51.92$_{5.34}$ & 71.86$_{0.65}$
    & 65.56$_{0.69}$ & 37.84$_{0.71}$ & 8.89$_{0.52}$ & 39.42$_{0.45}$ \\
    & mixup$^{\dag}$ & - 
    & \textbf{94.72}$_{0.43}$ & \textbf{75.70}$_{1.03}$ & 51.82$_{1.61}$ & 74.24$_{0.44}$
    & \textbf{70.43}$_{0.33}$ & 39.66$_{0.60}$ & 5.74$_{0.44}$ & \textbf{40.90}$_{0.25}$ \\
    & manifold mixup$^{\dag}$ & - 
    & 94.04$_{0.42}$ & 74.61$_{0.96}$ & 52.69$_{1.95}$ & 73.86$_{0.58}$
    & 69.63$_{0.49}$ & 38.83$_{0.93}$ & 5.23$_{0.60}$ & 40.18$_{0.25}$ \\
    & AM-mixup &  
    & 92.48$_{1.79}$ & 74.72$_{1.12}$ & \textbf{59.15}$_{3.45}$ & \textbf{75.38}$_{0.63}$
    & 66.04$_{0.65}$ & \textbf{40.60}$_{0.71}$ & \textbf{10.04}$_{0.45}$ & \textbf{40.90}$_{0.63}$ 
    \\
    \midrule
    \multirow{8}{*}{\rotatebox[origin=c]{90}{imb factor: 50}} 
    & CE & \cite{mislas_zhong2021improving}  
    & - & - & - & 74.80~~~~~~~ 
    & - & - & - & 43.90~~~~~~~ \\
    & mixup & \cite{mislas_zhong2021improving}
    & - & - & - & 77.80~~~~~~~ 
    & - & - & - & 45.00~~~~~~~  \\
    & manifold mixup  & \cite{bbn_zhou2020bbn} 
    & - & - & - & 77.95~~~~~~~ 
    & - & - & - & 43.09~~~~~~~ \\
    \cmidrule(lr){2-11}
    & CE$^{\dag}$ & -      
    & 93.30$_{1.26}$ & 76.74$_{1.00}$ & 62.98$_{1.91}$ & 77.58$_{0.48}$     
    & 66.80$_{0.75}$ & 44.51$_{0.83}$ & 16.71$_{0.87}$ & 44.48$_{0.55}$ \\
    & mixup$^{\dag}$ & -      
    & \textbf{94.98}$_{0.30}$ & \textbf{79.98}$_{0.70}$ & 65.04$_{1.58}$ & 80.00$_{0.54}$
    & \textbf{71.79}$_{0.48}$ & \textbf{46.92}$_{0.78}$ & 13.27$_{0.65}$ & \textbf{46.11}$_{0.36}$ \\
    & manifold mixup $^{\dag}$ & -      
    & 94.20$_{0.49}$ & 78.85$_{0.88}$ & 66.58$_{1.32}$ & 79.77$_{0.47}$
    & 71.13$_{0.73}$ & 45.63$_{0.59}$ & 12.04$_{0.93}$ & 45.07$_{0.39}$ \\
    & AM-mixup &       
    & 93.15$_{0.37}$ & 78.44$_{0.89}$ & \textbf{69.57}$_{1.27}$ & \textbf{80.19}$_{0.47}$ 
    & 66.84$_{0.64}$ & 46.45$_{0.69}$ & \textbf{17.56}$_{0.86}$ & 45.41$_{0.40}$
    \\
    \midrule
    \multirow{8}{*}{\rotatebox[origin=c]{90}{imb factor: 10}} 
    & CE & \cite{mislas_zhong2021improving}  
    & - & - & - & 86.40~~~~~~~ 
    & - & - & - & 55.80~~~~~~~ \\
    & mixup & \cite{mislas_zhong2021improving}
    & - & - & - & 87.10~~~~~~~ 
    & - & - & - & 58.20~~~~~~~  \\
    & manifold mixup  & \cite{bbn_zhou2020bbn} 
    & - & - & - & 87.03~~~~~~~ 
    & - & - & - & 56.55~~~~~~~ \\
    \cmidrule(lr){2-11}
    & CE$^{\dag}$ & -      
    & 94.00$_{0.40}$ & 84.77$_{0.45}$ & 84.37$_{0.64}$ & 87.42$_{0.30}$      
    & 68.90$_{0.60}$ & 57.71$_{0.75}$ & 41.63$_{0.59}$ & 57.07$_{0.39}$ \\
    & mixup$^{\dag}$ & -      
    & \textbf{95.37}$_{0.29}$ & \textbf{86.81}$_{0.47}$ & 85.80$_{0.89}$ & \textbf{89.08}$_{0.32}$
    & \textbf{73.86}$_{0.46}$ & \textbf{60.66}$_{0.54}$ & 41.08$_{0.71}$ & \textbf{59.73}$_{0.33}$ \\
    & manifold mixup $^{\dag}$ & -      
    & 94.62$_{0.32}$ & 86.25$_{0.39}$ & 86.06$_{0.58}$ & 88.71$_{0.21}$
    & 73.44$_{0.50}$ & 59.82$_{0.56}$ & 39.74$_{0.55}$ & 58.90$_{0.26}$ \\
    & AM-mixup &       
    & 93.47$_{0.47}$ & 85.26$_{0.58}$ & \textbf{86.82}$_{0.61}$ & 88.19$_{0.25}$ 
    & 69.25$_{0.54}$ & 59.00$_{0.58}$ & \textbf{42.91}$\textbf{}_{0.73}$ & 58.02$_{0.32}$ 
    \\
    \bottomrule
\end{tabular}
\label{tab:experiment_cifar_lt}
\end{table*}

\section{Experimental Results}

\subsection{Common Settings}
We chose two baseline methods for comparison with AM-mixup. 
Mixup effectively demonstrated our target problem, while manifold mixup served as a representative method for feature augmentation.
We conducted experiments with varying random seeds for imbalanced learning and coarse-to-fine transfer learning.
Their performances are represented by the mean \textit{mean} and standard deviation \textit{std}, i.e., \textit{mean}$_{\textit{std}}$. 
In AM-mixup, the default value of $\beta$ is set to 0.34, and we only provide annotation when it has a different value.
Mixup alpha is set to 0.2 in the ImageNet-LT and Places-LT, and 1.0 in all other datasets.
The hyperparameter settings are described in more detail in each experiment section.
In the following tables, $\dag$ denotes that the method was reproduced under the same experimental settings as our method in question.
We report accuracy across three class splits: \textit{Head-Many} (more than 100 images), \textit{Medium} (20 to 100 images), and \textit{Tail-Few} (less than 20 images).

\begin{table}[t]
\footnotesize
\centering
\caption{Results of Imbalanced Learning on ImageNet-LT. All experiments were conducted using 5 trials with random seeds. (Accuracy (\%): \textit{mean}$_{\textit{std}}$)}
\begin{tabular}{@{}llrrrr@{}}
    \toprule
     & \multirow{2}{*}{Method} & \multicolumn{4}{c}{ImageNet-LT} \\ 
    \cmidrule(ll){3-6}
    & & Many & Median & Few & All \\
    \midrule
    \multirow{4}{*}{\rotatebox[origin=c]{90}{ResNet50}}
    & CE$^{\dag}$ 
    & 66.53$_{0.18}$ & 40.34$_{0.45}$ & \textbf{12.03}$_{0.30}$ & \textbf{45.88}$_{0.31}$ \\
    & mixup$^{\dag}$ 
    & \textbf{67.40}$_{0.55}$ & 38.74$_{0.83}$ & 9.12$_{0.40}$ & 45.03$_{0.64}$ \\
    & manifold mixup$^{\dag}$ 
    & 67.27$_{0.42}$ & 39.01$_{0.87}$ & 9.65$_{0.53}$ & 45.19$_{0.63}$ \\[1pt]
    & AM-mixup   
    & 66.71$_{0.46}$ & \textbf{40.39}$_{0.61}$ & 10.34$_{0.18}$ & 45.70$_{0.45}$ \\
    \midrule
    \multirow{4}{*}{\rotatebox[origin=c]{90}{ResNet101}}
    & CE$^{\dag}$ 
    & 67.44$_{0.25}$ & 41.77$_{0.57}$ & \textbf{13.53}$_{0.54}$ & 47.12$_{0.41}$ \\
    & mixup$^{\dag}$ 
    & \textbf{69.31}$_{0.41}$ & 41.89$_{0.60}$ & 11.12$_{0.35}$ & 47.51$_{0.46}$ \\
    & manifold mixup$^{\dag}$ 
    & \textbf{69.31}$_{0.55}$ & 42.53$_{0.66}$ & 11.93$_{0.61}$ & \textbf{47.92}$_{0.60}$ \\[1pt]
    & AM-mixup$_{\beta=0.5}$   
    & 68.62$_{0.29}$ & \textbf{43.12}$_{0.03}$ & 11.02$_{0.25}$ & 47.77$_{0.15}$ \\
    \midrule
    \multirow{4}{*}{\rotatebox[origin=c]{90}{ResNet152}}
    & CE$^{\dag}$ 
    & 67.59$_{0.30}$ & 42.19$_{0.57}$ & \textbf{14.04}$_{0.30}$ & 47.45$_{0.38}$ \\
    & mixup$^{\dag}$  
    & 69.86$_{0.53}$ & 42.99$_{0.71}$ & 11.79$_{0.47}$ & 48.32$_{0.57}$ \\
    & manifold mixup$^{\dag}$ 
    & \textbf{69.93}$_{0.29}$ & \textbf{43.32}$_{0.29}$ & 12.43$_{0.37}$ & \textbf{48.59}$_{0.28}$ \\[1pt]
    & AM-mixup$_{\beta=0.2}$   
    & 67.80$_{0.06}$ & 42.69$_{0.47}$ & 12.68$_{0.63}$ & 47.53$_{0.34}$ \\
    \bottomrule
\end{tabular}
\label{tab:experiment_imagenet_lt}
\end{table}

\begin{table}[t]
\footnotesize
\centering
\caption{Results of Imbalanced Learning on Places-LT. All experiments were conducted using 3 trials with random seeds. (Accuracy (\%): \textit{mean}$_{\textit{std}}$)}
\begin{tabular}{@{}lrrrr@{}}
    \toprule
    \multirow{2}{*}{Method} & \multicolumn{4}{c}{Places-LT} \\ 
    \cmidrule(ll){2-5}
    & Many & Median & Few & All \\
    \midrule
    CE$^{\dag}$ 
    & 40.63$_{0.21}$ & 17.69$_{0.64}$ & 2.04$_{0.44}$ & 22.62$_{0.29}$ \\
    mixup$^{\dag}$ 
    & 42.10$_{0.31}$ & 15.82$_{0.60}$ & 0.86$_{0.18}$ & 22.10$_{0.27}$ \\
    manifold mixup$^{\dag}$ 
    & \textbf{42.50}$_{0.11}$ & 16.87$_{0.71}$ & 1.15$_{0.21}$ & 22.76$_{0.31}$ \\[0.8pt]
    AM-mixup$_{\beta=0.1}$   
    & 41.15$_{0.35}$ & \textbf{18.76}$_{0.33}$ & \textbf{2.21}$_{0.09}$ & \textbf{23.31}$_{0.05}$ \\
    \bottomrule
\end{tabular}
\label{tab:experiment_places_lt}
\end{table}

\subsection{Inter-class Collapse in Imbalanced Learning}
\label{subsec:imbalanced_learning}


\paragraph{Result and Analysis}
As shown in \autoref{tab:experiment_cifar_lt}, AM-mixup attains the best performance in CIFAR10/100-LT, except for imbalance factors 50 and 10 on CIFAR100-LT.
Mixup shows better performance than CE but not as good as AM-mixup.
This implies that mixup suffers from inter-class collapse more than AM-mixup in the long-tailed datasets.
The improvement of performance in mixup and manifold mixup was primarily achieved by increasing precision in the \textit{many} and \textit{median} parts.
It means that they do not obtain enough margin for the tail classes, whereas AM-mixup achieves better margin balance by improving the precision of the \textit{few} part.
As a result, AM-mixup achieves the highest performance by reducing inter-class collapse between tail classes.
However, this tendency differs slightly on more empirical dataset, ImageNet-LT.
In the ImageNet-LT dataset, CE exhibits better precision than interpolation-based methods including AM-mixup in the \textit{few} part, and even in the case of ResNet50, CE shows better performance than the others, as illustrated in~\autoref{tab:experiment_imagenet_lt}.
Despite this inferiority, AM-mixup outperforms mixup and manifold mixup on the \textit{few} part in ResNet50 and ResNet152, while showing comparable performance in ResNet101.
This implies that asymptotic movement is beneficial for alleviating the collapse between tail classes.
In contrast, the result of experiments on Places-LT shows a similar tendency to that in CIFAR10/100-LT, as demonstrated in~\autoref{tab:experiment_places_lt}.

\begin{table}[t]
\footnotesize
  \centering
  \caption{Coarse-to-Fine Transfer Learning on CIFAR10/100. All experiments were conducted using 5 trials with random seeds. Best in bold and second best in underline. (Accuracy (\%): \textit{mean}$_{\textit{std}}$)}
  \begin{tabular}{@{}lcccc@{}}
    \toprule
    Method & ref & MNIST & CIFAR10 & CIFAR100 \\
    \midrule
    CE & \cite{contrast_chen2022perfectly} & 98.70$_{0.10}$ & 71.10$_{0.20}$ & 54.20$_{0.20}$ \\
    \midrule
    CE$^{\dag}$ & - 
    & \underline{98.12}$_{0.15}$ & \textbf{70.73}$_{0.27}$ & \textbf{59.97}$_{0.36}$ \\
    mixup & - 
    & 97.07$_{0.46}$ & 64.16$_{0.75}$ & 59.32$_{0.25}$ \\
    AM-mixup &  
    & \textbf{98.20}$_{0.12}$ & \underline{66.80}$_{0.56}$ & \underline{59.42}$_{0.21}$ \\
    \bottomrule
  \end{tabular}
  \label{tab:experiment_coarse_to_fine}
\end{table}

\subsection{Intra-class Collapse in Coarse-to-Fine Transfer Learning}

\paragraph{Result and Analysis.}
As shown in Table~\ref{tab:experiment_coarse_to_fine}, AM-mixup achieves the second-best test accuracy, and even outperformed all others on MNIST.
However, mixup exhibits intra-class collapse, as indicated by its low accuracy. 
Similarly, AM-mixup also has intra-class collapse by showing lower accuracy than the cross entropy without any augmentation, but AM-mixup achieves better than mixup.
It means that the mixup rate control and one-sided labeling of AM-mixup alleviate intra-class collapse.

\begin{table}[t]
\footnotesize
\centering
\caption{Ablation study on CIFAR100-LT. MR: Mixup Rate, OL: one-sided Labeling, LL: Last Layer, $\lambda$: mixup rate randomly selected from beta distribution. $\lambda_{0.51}$: fixed mixup rate as 0.51, $\lambda_{am}$: AM-mixup rate. All experiments were conducted using 5 trials with random seeds. Best in bold. (Accuracy (\%): \textit{mean}$_{\textit{std}}$)}
\begin{tabular}{@{}cccrrrr@{}}
    \toprule
    \multirow{2}{*}{MR} & 
    \multirow{2}{*}{OL} &
    \multirow{2}{*}{LL} &
    \multicolumn{4}{c}{CIFAR100-LT (imb: 100)} \\ 
    \cmidrule(ll){4-7}
    & & & Many & Median & Few & All \\
    \midrule
    $\lambda_{am}$ &  & 
    & \textbf{68.77}$_{0.29}$ & 37.86$_{0.86}$ & 6.28$_{0.56}$ & 39.83$_{0.36}$ \\
    $\lambda_{am}$ &  & \checkmark
    & 65.63$_{0.75}$ & 37.67$_{0.94}$ & 8.34$_{0.71}$ & 39.23$_{0.36}$ \\
    $\lambda_{am}$ & \checkmark & 
    & 67.15$_{0.49}$ & 40.95$_{0.45}$ & 9.23$_{0.45}$ & \textbf{41.19}$_{0.28}$ \\
    $\lambda $ & \checkmark & \checkmark
    & 66.97$_{0.79}$ & 36.68$_{1.14}$ & 2.91$_{0.34}$ & 37.79$_{0.56}$ \\
    $\lambda_{0.51} $ & \checkmark & \checkmark
    & 66.84$_{0.67}$ & 39.73$_{0.95}$ & 6.57$_{0.29}$ & 39.87$_{0.17}$ \\
    $\lambda_{am} $ & \checkmark & \checkmark
    & 65.78$_{0.55}$ & \textbf{40.97}$_{0.47}$ & \textbf{10.03}$_{0.27}$ & 40.93$_{0.29}$ \\
    \bottomrule
\end{tabular}
\label{tab:experiment_ablation_only_cifar100}
\end{table}

\subsection{Ablation Study}
To qualitatively compare the effects of main compoments in AM-mixup, we conducted the ablation study on CIFAR100-LT with the imbalance factor set to 100.
We used the same experimental and hyperparameter settings to~\autoref{subsec:imbalanced_learning}.
\textit{Effect of Mixup Rate} (rows 4-6 of~\autoref{tab:experiment_ablation_only_cifar100}).
As shown in~\autoref{tab:experiment_ablation_only_cifar100}, the model's performance varies depending on the scheduling of the mixup rate.
The use of $\lambda_{am}$ results in the best performance.
As a result, it is important to adjust the mixup rate based on the model's reliability.
\textit{Effect of One-sided Labeling} (rows 1-3, 6 of~\autoref{tab:experiment_ablation_only_cifar100}).
Checkmark in the column of OL means that the model used the one-sided labels as the ground truth labels in training.
As illustrated in~\autoref{tab:experiment_ablation_only_cifar100}, when comparing the performance of models under equal conditions (rows 1, 3 and rows 2, 6), the use of one-sided labels consistently results in superior performance than not using them.
Consequently, it is beneficial to consider the features augmented from the adjusted mixup rate by AM-mixup as one-sided features, which helps in training more confident models when training datasets are imbalanced.
\textit{Effect of Mixup only at the Last Layer} (rows 1-3, 6 of~\autoref{tab:experiment_ablation_only_cifar100}).
Checkmark in the column of LL indicates that the model applied mixup only to the features of the last layer.
As shown in~\autoref{tab:experiment_ablation_only_cifar100}, when comparing the performance of models under identical conditions (rows 1, 2 and rows 3, 6), LL exhibits worse performance.
Nonetheless, it outperforms in the median and few parts, which are critical points in imbalanced learning.
As a result, applying mixup only at the last layer aids in balancing the margin by estimating the decision boundary and augmenting features accordingly.

%
%
\section{Related Work}
\label{related_work}

\subsection{Augmentation}
Data augmentation has proven to be an effective regularization technique~\cite{mixup, rw_shorten2019survey, rw_devries2017improved, rw_cubuk2018autoaugment, rw_zhong2020random, rw_moreno2018forward}.
Mixup~\cite{mixup}, a commonly used approach in data augmentation, interpolates each pair of input samples and labels in the input space, which allows models to enhance their inductive bias. 
In other streams, data augmentation has been applied to features in the feature space, also known as feature augmentation~\cite{manifold_mixup, rw_li2021simple, rw_kuo2020featmatch, rw_lee2021learning, rw_wang2021regularizing}.
In manifold mixup~\cite{manifold_mixup}, models obtain a smoother decision boundary, leading to better robustness. 
However, the emphasis is not placed on the margin, a critical element for creating a robust decision boundary.
In contrast, our proposed method generates augmented features within the feature space and adjusts the augmentation to balance and moderately widen the margin.

\subsection{Contrastive Learning}
Contrastive learning has accomplished state-of-the-art performance in image classification tasks, exemplifying a focus on margin~\cite{contrast_chen2020simple, contrast_he2020momentum, contrast_caron2020unsupervised, contrast_li2020prototypical, contrast_gutmann2010noise, contrast_koch2015siamese, contrast_supcon_khosla2020supervised}. 
This approach draws positive samples towards the anchor while repelling negative samples. 
In supervised methods, label information is used by SupCon~\cite{contrast_supcon_khosla2020supervised} to select both positive and negative pairs.
As a result, SupCon achieves a high level of uniformity between inter-class and minor alignment between intra-class. 
This property results in ideal representations with a generous margin between classes. 
However, SupCon suffers from the issue of \textit{collapse}~\cite{contrast_jing2021understanding} where each sample converges towards the class centroid. 
This leads to feature becoming indistinguishable from each other, resulting in subpar performance during coarse-to-fine transfer learning~\cite{contrast_chen2022perfectly}. 
Furthermore, previous studies have shown relatively low performance in imbalanced learning when using SupCon~\cite{contrast_zhu2022balanced, contrast_li2022targeted}. 
In imbalanced learning, the use of SupCon can result in an excessive focus on head classes, potentially leading to the merging of tail classes.
To address this, BCL~\cite{contrast_zhu2022balanced} employs SupCon loss with class-average and class-complement, while TSC~\cite{contrast_li2022targeted} enforces class centroids to form a regular simplex on the hypersphere. 
In comparison, our approach avoids collapse by generating augmented features that asymptotically move towards the midpoint, resulting in a balanced and moderately broad margin.

%
%
\section{Conclusion}
\label{conclusion}
In this paper, we address the two collapse problems surrounding interpolation-based augmentation methods, which have recently been discussed in contrastive literature.
Through analyzing alignment and uniformity, used as indicators of collapse problems, we have found that such problems remain significant even in famous interpolation-based augmentation methods such as mixup and manifold mixup.
To address the issues of collapse, we propose the \textit{Asymptotic Midpoint mixup} method to generate effective features through interpolation of features with one-sided labeling and their asymptotic movement towards the midpoint.
The method demonstrates the effects of margin balancing and moderate-broadening, and their impact on the problems of collapse through quantitative and qualitative analysis of a toy dataset.
In practical experiments involving imbalanced learning and coarse-to-fine transfer learning, which encountered issues with inter-class and intra-class collapse, our method provided significantly greater relief to the problem than mixup and manifold mixup.
However, it must be noted that AM-mixup may require additional tuning of the hyperparameter $\beta$ to achieve the best performance, given the varying intensities of collapse problems across tasks.

{
\small

\bibliography{reference}

\begin{thebibliography}{10}

\bibitem{manifold_mixup}
Vikas Verma, Alex Lamb, Christopher Beckham, Amir Najafi, Ioannis Mitliagkas, David Lopez-Paz, and Yoshua Bengio.
\newblock Manifold mixup: Better representations by interpolating hidden states.
\newblock In {\em International conference on machine learning}, pages 6438--6447. PMLR, 2019.

\bibitem{contrast_chen2020simple}
Ting Chen, Simon Kornblith, Mohammad Norouzi, and Geoffrey Hinton.
\newblock A simple framework for contrastive learning of visual representations.
\newblock In {\em International conference on machine learning}, pages 1597--1607. PMLR, 2020.

\bibitem{contrast_he2020momentum}
Kaiming He, Haoqi Fan, Yuxin Wu, Saining Xie, and Ross Girshick.
\newblock Momentum contrast for unsupervised visual representation learning.
\newblock In {\em Proceedings of the IEEE/CVF conference on computer vision and pattern recognition}, pages 9729--9738, 2020.

\bibitem{contrast_li2022targeted}
Tianhong Li, Peng Cao, Yuan Yuan, Lijie Fan, Yuzhe Yang, Rogerio~S Feris, Piotr Indyk, and Dina Katabi.
\newblock Targeted supervised contrastive learning for long-tailed recognition.
\newblock In {\em Proceedings of the IEEE/CVF Conference on Computer Vision and Pattern Recognition}, pages 6918--6928, 2022.

\bibitem{contrast_chen2022perfectly}
Mayee Chen, Daniel~Y Fu, Avanika Narayan, Michael Zhang, Zhao Song, Kayvon Fatahalian, and Christopher R{\'e}.
\newblock Perfectly balanced: Improving transfer and robustness of supervised contrastive learning.
\newblock In {\em International Conference on Machine Learning}, pages 3090--3122. PMLR, 2022.

\bibitem{contrast_supcon_khosla2020supervised}
Prannay Khosla, Piotr Teterwak, Chen Wang, Aaron Sarna, Yonglong Tian, Phillip Isola, Aaron Maschinot, Ce~Liu, and Dilip Krishnan.
\newblock Supervised contrastive learning.
\newblock {\em Advances in neural information processing systems}, 33:18661--18673, 2020.

\bibitem{mixup}
Hongyi Zhang, Moustapha Cisse, Yann~N Dauphin, and David Lopez-Paz.
\newblock mixup: Beyond empirical risk minimization.
\newblock {\em arXiv preprint arXiv:1710.09412}, 2017.

\bibitem{mixup_ssl}
Yichen Zhang, Yifang Yin, Ying Zhang, and Roger Zimmermann.
\newblock Mix-up self-supervised learning for contrast-agnostic applications, 2022.

\bibitem{measure_wang2020understanding}
Tongzhou Wang and Phillip Isola.
\newblock Understanding contrastive representation learning through alignment and uniformity on the hypersphere.
\newblock In {\em International Conference on Machine Learning}, pages 9929--9939. PMLR, 2020.

\bibitem{mislas_zhong2021improving}
Zhisheng Zhong, Jiequan Cui, Shu Liu, and Jiaya Jia.
\newblock Improving calibration for long-tailed recognition.
\newblock In {\em Proceedings of the IEEE/CVF conference on computer vision and pattern recognition}, pages 16489--16498, 2021.

\bibitem{bbn_zhou2020bbn}
Boyan Zhou, Quan Cui, Xiu-Shen Wei, and Zhao-Min Chen.
\newblock Bbn: Bilateral-branch network with cumulative learning for long-tailed visual recognition.
\newblock In {\em Proceedings of the IEEE/CVF conference on computer vision and pattern recognition}, pages 9719--9728, 2020.

\bibitem{rw_shorten2019survey}
Connor Shorten and Taghi~M Khoshgoftaar.
\newblock A survey on image data augmentation for deep learning.
\newblock {\em Journal of big data}, 6(1):1--48, 2019.

\bibitem{rw_devries2017improved}
Terrance DeVries and Graham~W Taylor.
\newblock Improved regularization of convolutional neural networks with cutout.
\newblock {\em arXiv preprint arXiv:1708.04552}, 2017.

\bibitem{rw_cubuk2018autoaugment}
Ekin~D Cubuk, Barret Zoph, Dandelion Mane, Vijay Vasudevan, and Quoc~V Le.
\newblock Autoaugment: Learning augmentation policies from data.
\newblock {\em arXiv preprint arXiv:1805.09501}, 2018.

\bibitem{rw_zhong2020random}
Zhun Zhong, Liang Zheng, Guoliang Kang, Shaozi Li, and Yi~Yang.
\newblock Random erasing data augmentation.
\newblock In {\em Proceedings of the AAAI conference on artificial intelligence}, volume~34, pages 13001--13008, 2020.

\bibitem{rw_moreno2018forward}
Francisco~J Moreno-Barea, Fiammetta Strazzera, Jos{\'e}~M Jerez, Daniel Urda, and Leonardo Franco.
\newblock Forward noise adjustment scheme for data augmentation.
\newblock In {\em 2018 IEEE symposium series on computational intelligence (SSCI)}, pages 728--734. IEEE, 2018.

\bibitem{rw_li2021simple}
Pan Li, Da~Li, Wei Li, Shaogang Gong, Yanwei Fu, and Timothy~M Hospedales.
\newblock A simple feature augmentation for domain generalization.
\newblock In {\em Proceedings of the IEEE/CVF International Conference on Computer Vision}, pages 8886--8895, 2021.

\bibitem{rw_kuo2020featmatch}
Chia-Wen Kuo, Chih-Yao Ma, Jia-Bin Huang, and Zsolt Kira.
\newblock Featmatch: Feature-based augmentation for semi-supervised learning.
\newblock In {\em Computer Vision--ECCV 2020: 16th European Conference, Glasgow, UK, August 23--28, 2020, Proceedings, Part XVIII 16}, pages 479--495. Springer, 2020.

\bibitem{rw_lee2021learning}
Jungsoo Lee, Eungyeup Kim, Juyoung Lee, Jihyeon Lee, and Jaegul Choo.
\newblock Learning debiased representation via disentangled feature augmentation.
\newblock {\em Advances in Neural Information Processing Systems}, 34:25123--25133, 2021.

\bibitem{rw_wang2021regularizing}
Yulin Wang, Gao Huang, Shiji Song, Xuran Pan, Yitong Xia, and Cheng Wu.
\newblock Regularizing deep networks with semantic data augmentation.
\newblock {\em IEEE Transactions on Pattern Analysis and Machine Intelligence}, 44(7):3733--3748, 2021.

\bibitem{contrast_caron2020unsupervised}
Mathilde Caron, Ishan Misra, Julien Mairal, Priya Goyal, Piotr Bojanowski, and Armand Joulin.
\newblock Unsupervised learning of visual features by contrasting cluster assignments.
\newblock {\em Advances in neural information processing systems}, 33:9912--9924, 2020.

\bibitem{contrast_li2020prototypical}
Junnan Li, Pan Zhou, Caiming Xiong, and Steven~CH Hoi.
\newblock Prototypical contrastive learning of unsupervised representations.
\newblock {\em arXiv preprint arXiv:2005.04966}, 2020.

\bibitem{contrast_gutmann2010noise}
Michael Gutmann and Aapo Hyv{\"a}rinen.
\newblock Noise-contrastive estimation: A new estimation principle for unnormalized statistical models.
\newblock In {\em Proceedings of the thirteenth international conference on artificial intelligence and statistics}, pages 297--304. JMLR Workshop and Conference Proceedings, 2010.

\bibitem{contrast_koch2015siamese}
Gregory Koch, Richard Zemel, Ruslan Salakhutdinov, et~al.
\newblock Siamese neural networks for one-shot image recognition.
\newblock In {\em ICML deep learning workshop}, volume~2. Lille, 2015.

\bibitem{contrast_jing2021understanding}
Li~Jing, Pascal Vincent, Yann LeCun, and Yuandong Tian.
\newblock Understanding dimensional collapse in contrastive self-supervised learning.
\newblock {\em arXiv preprint arXiv:2110.09348}, 2021.

\bibitem{contrast_zhu2022balanced}
Jianggang Zhu, Zheng Wang, Jingjing Chen, Yi-Ping~Phoebe Chen, and Yu-Gang Jiang.
\newblock Balanced contrastive learning for long-tailed visual recognition.
\newblock In {\em Proceedings of the IEEE/CVF Conference on Computer Vision and Pattern Recognition}, pages 6908--6917, 2022.

\bibitem{arch_he2016deep}
Kaiming He, Xiangyu Zhang, Shaoqing Ren, and Jian Sun.
\newblock Deep residual learning for image recognition.
\newblock In {\em Proceedings of the IEEE conference on computer vision and pattern recognition}, pages 770--778, 2016.

\bibitem{vit_dosovitskiy2020image}
Alexey Dosovitskiy, Lucas Beyer, Alexander Kolesnikov, Dirk Weissenborn, Xiaohua Zhai, Thomas Unterthiner, Mostafa Dehghani, Matthias Minderer, Georg Heigold, Sylvain Gelly, et~al.
\newblock An image is worth 16x16 words: Transformers for image recognition at scale.
\newblock {\em arXiv preprint arXiv:2010.11929}, 2020.

\bibitem{arch_simonyan2014very}
Karen Simonyan and Andrew Zisserman.
\newblock Very deep convolutional networks for large-scale image recognition.
\newblock {\em arXiv preprint arXiv:1409.1556}, 2014.

\bibitem{arch_huang2017densely}
Gao Huang, Zhuang Liu, Laurens Van Der~Maaten, and Kilian~Q Weinberger.
\newblock Densely connected convolutional networks.
\newblock In {\em Proceedings of the IEEE conference on computer vision and pattern recognition}, pages 4700--4708, 2017.

\end{thebibliography}
\bibliographystyle{unsrt}

}

\newpage

\appendix

\section{Additional Implementation Details.}
We describe details about the datasets, model architectures, and hyperparameter settings in the motivation, coarse-to-fine transfer learning, imbalanced learning and image classification benchmarks.

\subsection{Motivation}
\label{sec:appx_motivation}

\paragraph{Dataset.}

\begin{itemize}
    \item \textbf{Mini-CIFAR} consists of four classes: airplane, automobile, bird, cat. As a result, there are 20K samples total, 5K each per class in the dataset.
    \item \textbf{Mini-CIFAR-Coarse} consists of two superclasses: vehicles(airplane, automobile) and animals(bird, cat)
    \item \textbf{Mini-CIFAR-LT} consists of four imbalanced classes. According to the general settings in long-tailed datasets, we subsamples from Mini-CIFAR at an exponentially decreasing rate from the first. 
    The imbalance factor means the ratio of the number of a head class $n_{max}$ to the number of a tail class $n_{min}$, i.e., $imb=\frac{n_{max}}{n_{min}}$. 
    We set the imbalance factor as 200, and as a result, the number of samples in each class is \{\textit{airplane}: 5000, \textit{automobile}: 854, \textit{bird}: 146, \textit{cat}: 25\}.
\end{itemize}

\paragraph{Model Architectures.}
We utilized CNN-Vis2D to visualize the results on 2-dimensional space. 
CNN-Vis2D is composed of three convolutional layer blocks whose kernel size are 32, 64, and 128.
To obtain 2-dimensional feature vectors, we set the input dimension of the last layer as 2.

\paragraph{Hyperparameters.}
We trained the model with the softmax function, cross entropy, and SGD with momentum 0.9 and weight decay 5e-4. 
The initial learning rate was set to 0.1 and divided by 0.2 at epoch 30, 60, and 80 during 100 epochs. 

\subsection{Imbalanced Learning}
\label{appx:subsec_imbalanced_learning}
We borrowed the same settings as~\cite{mislas_zhong2021improving} in both datasets and model architectures

\paragraph{Datasets.}
\begin{itemize}
    \item \textbf{CIFAR10-LT} consists of ten imbalanced classes. 
    In the same way we already mentioned in \autoref{sec:appx_motivation}, CIFAR10-LT was constructed by subsampling from CIFAR10 at an exponentially decreasing rate from the first class.
    \item \textbf{CIFAR100-LT} consists of one hundred imbalanced classes.
    In the same way we already mentioned in \autoref{sec:appx_motivation}, CIFAR100-LT was constructed by subsampling from CIFAR100 at an exponentially decreasing rate from the first class.    
    \item \textbf{ImageNet-LT} is a long-tailed version of the large-scale object classification dataset ImageNet.
    It contains 115.8K images, distributed across 1,000 classes, by sampling a subset following the Pareto distribution with a power value $\alpha$=5.
    The classes have varying class cardinality, ranging from 5 to 1,280.
    \item \textbf{Places-LT} is a long-tailed version of the large-scale scene classification dataset Places.
    It contains 184.5K images from 365 classes.
    The classes differ in their cardinality, ranging from 5 to 4,980.
\end{itemize}

\paragraph{Model Architectures.}
We use a ResNet32~\cite{mislas_zhong2021improving} (3 residual blocks, each output dimension is 16, 32, and 64) for CIFAR10/100-LT datasets. 
Different with ResNet architecture for ImageNet, the kernel size, stride, and padding are set to 3, 1, and 1, respectly at the first convolutional layer.
ResNet50, 101, and 152 are same to ~\cite{arch_he2016deep}.

\paragraph{Hyperparameters.}
We used ResNet32 on CIFAR10/100-LT and trained the model with 128 mini-batches, SGD at the momentum of 0.9 and weight decay of 2e-4, and the number of epochs is 200.
We warmed up the learning rate from 0.02 to the initial learning rate linearly and it was divided by 0.1 at epoch 160 and 180.
For the other datasets, we utilized ResNet50, 101, and 152, trained the models with SGD at the momentum of 0.9 and weight decay of 5e-4, and set the learning rate using a cosine annealing schedule.
We also set mixup alpha as 1.0 ($\alpha$=1) for mixup and AM-mixup.

\subsection{Coarse-to-Fine Transfer Learning}
\label{appx:subsec_coarse_to_fine}
We borrowed the same settings as~\cite{contrast_chen2022perfectly} in both datasets and model architectures.

\paragraph{Datasets.}
\begin{itemize}
    \item \textbf{MNIST, CIFAR10}, and \textbf{CIFAR100} are all the standard image classification datasets.
    \item \textbf{MNIST-Coarse} consists of 2 superclasses: $<$5 (0, 1, 2, 3, 4) and $\geq$5 (5, 6, 7, 8, 9).
    \item \textbf{CIFAR10-Coarse} consists of 2 superclasses: animals (dog, cat, deer, horse, frog, bird) and vehicles (car, truck, plane, boat).
    \item \textbf{CIFAR100-Coarse} consists of 20 superclasses\footnote{https://www.cs.toronto.edu/~kriz/cifar.html}.
\end{itemize}

\paragraph{Model Architectures.}
We use a ViT model~\cite{vit_dosovitskiy2020image} (4 x 4 patch size, 7 multi-head attention layers with 8 attention heads and hidden MLP size of 256, final embedding size of 128) as the encoder. 

\paragraph{Hyperparameters.}
For the coarse dataset training, all models were trained for 600 epochs with an initial learning rate of 3e-4, a cosine annealing learning rate scheduler with $T_{max}$=100 and the AdamW optimizer with no weight decay. 
A dropout rate of 0.05 was used.
In the coarse-to-fine transfer experiments, all models are trained for 100 epochs with an initial learning rate of 1e-3, a cosine annealing learning rate scheduler with $T_{max}$=100 and the AdamW optimizer with no weight decay.
All experiments were run using a batch size of 128 for both training and evaluation.

\section{Additional Experiments}

\subsection{General Impact in image classification benchmarks}

\paragraph{Datasets.} 
For the image classification benchmarks, we employed CIFAR10, CIFAR100 and Tiny-ImageNet. 

\paragraph{Model Architectures.}
VGG11~\cite{arch_simonyan2014very}, ResNet50~\cite{arch_he2016deep} and DenseNet-BC with 12 growth rate~\cite{arch_huang2017densely} are used for the image classification benchmarks. 

\paragraph{Implementation Details.}
We conducted image classification experiments on CIFAR10, CIFAR100, and Tiny-ImageNet. For CIFAR10, we set the initial learning rate as 0.05 and divided the learning rate by two at every 30 epochs among the total of 300 epochs for all networks. For CIFAR100, For CIFAR100, we set the initial learning rate as 0.1 and divided it by five at the 60th, 120th, and 160th epochs, where the total number of epochs is 200 for all networks. For Tiny-ImageNet on VGG11 and ResNet50, we used 256 mini-batches, the SGD at a momentum of 0.9 without weight decay, and the number of epochs is 200. We set the initial learning rate as 0.1 and multiplied it by 0.9 at every 20 epochs. For DenseNet-BC ($k=12$) on Tiny-ImageNet, we used 64 mini-batches, the SGD at a momentum of 0.9 without weight decay, and the number of epochs is 300. We set the initial learning rate as 0.1 and divided it by ten at 150 and 225 epochs. 

\paragraph{Result and Analysis.}
As shown in Table~\ref{tab:image_classification_benchmark}, AM-mixup achieved competitive or even higher performance than cross entropy withou any augmentation. 
It implies AM-mixup sustains proper alignment and high uniformity without interruption for representation learning.

\begin{table}[t]
\centering
\caption{Performance in Image Classification Benchmarks. In AM-mixup, $\beta$ was set to 0.67. Best in bold (Accuracy (\%): \textit{mean}$_{\textit{std}}$)}
\begin{tabular}{clrrr}
\toprule
Network & Method & CIFAR-10 & CIFAR-100 & Tiny-ImageNet \\
\toprule
\multirow{4}{*}{VGG11} 
& CE        & 91.77$_{0.02}$ & 68.27$_{0.19}$ & 53.10$_{0.09}$ \\
& mixup     & \textbf{92.81$_{0.23}$} & \textbf{70.58$_{0.07}$} & \textbf{54.67$_{0.33}$} \\
& M-mixup   & 92.14$_{0.19}$ & 69.38$_{0.18}$ & 52.42$_{0.29}$ \\
\cmidrule{2-5}
& AM-mixup  & 92.68$_{0.20}$ & 70.49$_{0.15}$ & 54.48$_{0.33}$ \\
\midrule
\multirow{4}{*}{ResNet50} 
& CE        & 95.01$_{0.12}$ & 76.56$_{0.18}$ & 56.85$_{0.74}$ \\
& mixup     & \textbf{96.21$_{0.05}$} & \textbf{78.37$_{0.47}$} & \textbf{59.11$_{0.90}$} \\
& M-mixup   & 94.55$_{0.39}$ & 76.42$_{0.69}$ & 57.83$_{1.63}$ \\
\cmidrule{2-5}
& AM-mixup  & 95.38$_{0.07}$ & 77.05$_{0.65}$ & 58.36$_{0.16}$ \\
\midrule
\multirow{4}{*}{DenseNet-BC}  
& CE        & 94.92$_{0.02}$ & 76.96$_{0.27}$ & 60.23$_{0.34}$ \\
& mixup     & \textbf{95.54$_{0.09}$} & \textbf{77.23$_{0.28}$} & \textbf{62.93$_{0.10}$} \\
& M-mixup   & 94.57$_{0.18}$ & 76.54$_{0.11}$ & 62.43$_{0.09}$ \\
\cmidrule{2-5}
& AM-mixup  & 94.64$_{0.18}$ & 76.93$_{0.12}$ & 61.64$_{0.42}$ \\
\bottomrule
\end{tabular}
\label{tab:image_classification_benchmark}
\end{table}

\section{Additional Experimental Results}
\begin{figure*}[t]
    \centering
    \includegraphics[width=0.9\linewidth]{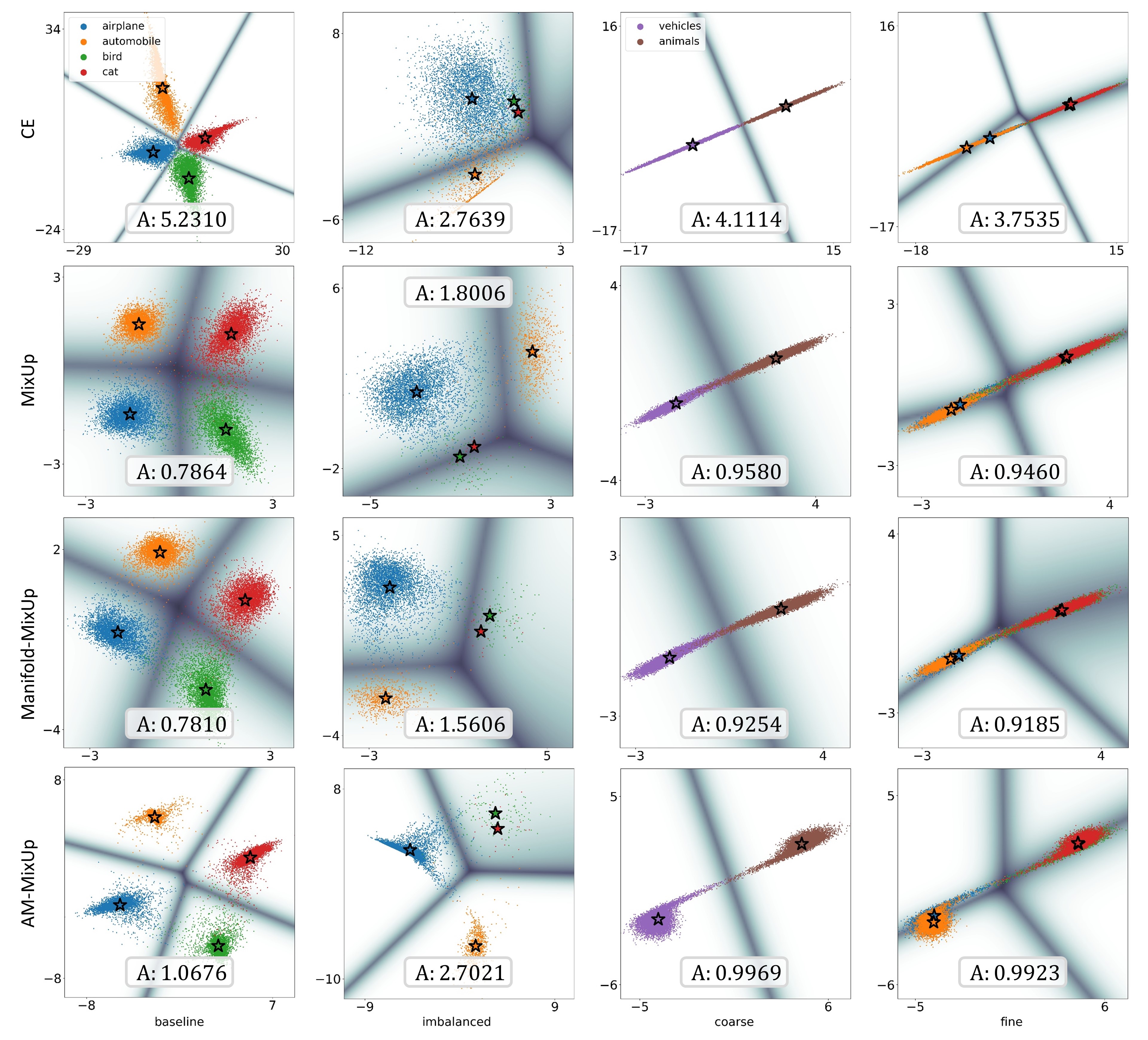}
    \caption{
    Visualization of the results of motivation test. Each alignment is printed on the figures.
    }
    \label{fig:toy_cartesian_detail}
\end{figure*}

\begin{figure*}[t]
    \centering
    \includegraphics[width=0.9\linewidth]{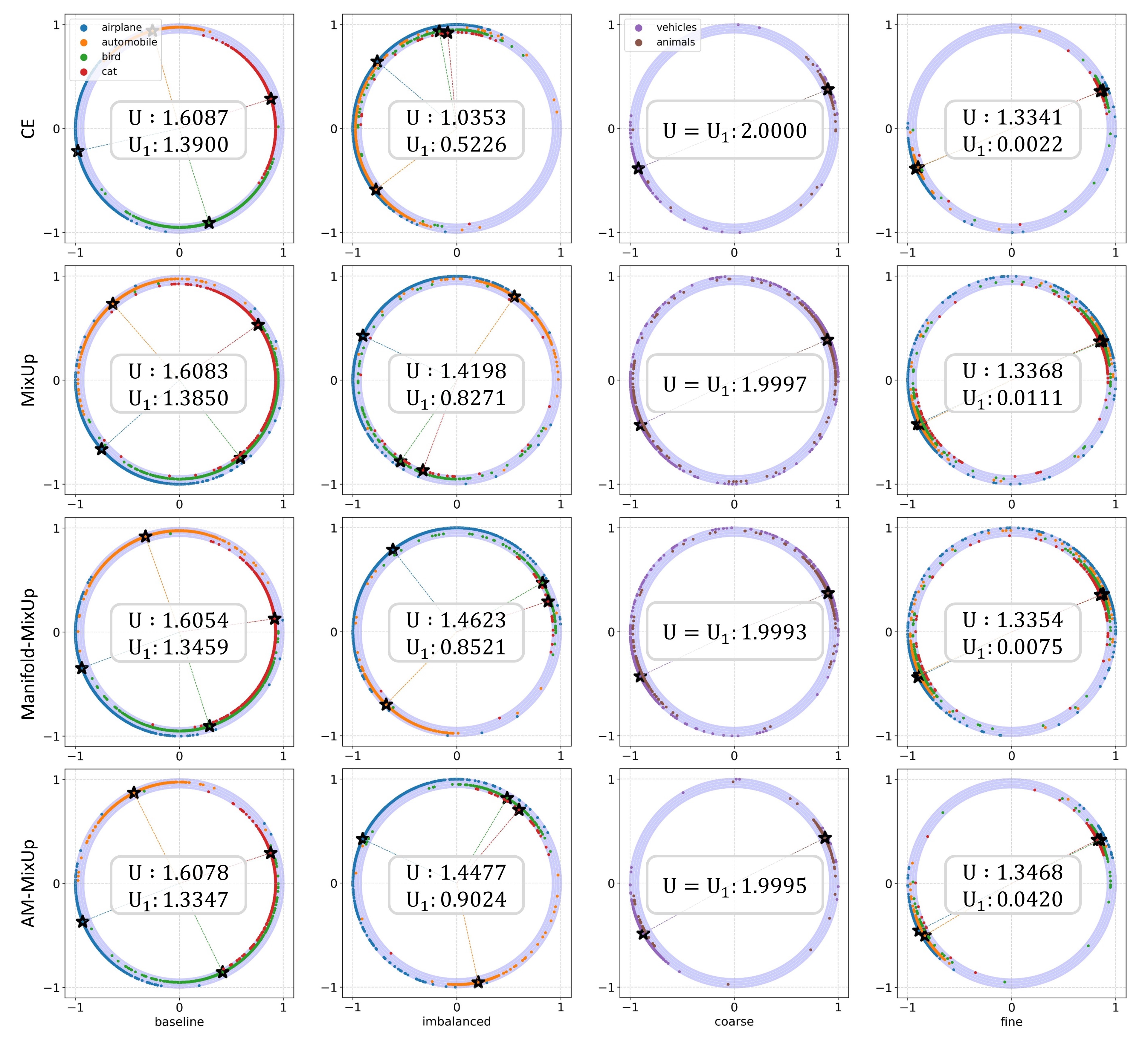}
    \caption{
    Visualization of the results of motivation test. Each uniformity and neighbor uniformity are printed on the figures.
    }
    \label{fig:toy_norm_detail}
\end{figure*}

We present additional experimental results about \autoref{sec:motivation}.
To make it easy to compare at a glance, we print the alignment $\textbf{A}$, inter-class uniformity $\textbf{U}$, and neighborhood($k$=1) uniformity $\textbf{U}_1$ on the figures, as shown \autoref{fig:toy_cartesian_detail}.
Additionally, we also give the figures that scatter features on the circle in 2-dimensional space like the last three rows in \autoref{fig:toy_norm_detail} to better synchronize $\textbf{U}$ and $\textbf{U}_1$ with features.

\end{document}